\documentclass[a4paper, 10pt, conference]{IEEEtran}

\IEEEoverridecommandlockouts 

\usepackage[english]{babel}
\usepackage{times} 
\usepackage{cite}
\usepackage[utf8]{inputenc}
\usepackage{hyperref}
\usepackage[nolist]{acronym}

\makeatletter
\g@addto@macro{\UrlBreaks}{\UrlOrds}
\makeatother

\usepackage{siunitx}
\usepackage{graphicx} 
\usepackage{epsfig} 
\usepackage{pgfplots}
\usepackage{caption}
\usepackage{subcaption}

\usepackage[export]{adjustbox}

\usepackage{mathtools}
\usepackage{nicefrac}
\usepackage{amsmath}
\usepackage{amssymb}
\usepackage{bm} 

\DeclareMathOperator*{\minimize}{minimize}

\DeclareMathOperator*{\argmin}{arg\,min}

\usepackage{tikz}
\pgfdeclarelayer{foreground}
\pgfsetlayers{background, main, foreground}
\usetikzlibrary{quotes, angles, backgrounds, arrows, automata, shapes, positioning, calc, through, spy, decorations.pathreplacing, decorations.markings, arrows.meta, automata, petri}

\tikzset{
    imglabel/.style={
      rectangle,
      inner sep=2pt,
      text=black,
      minimum height=1em,
      text centered,
      fill=white,
      fill opacity=1.0,
      text opacity=1,
      anchor=south west,
    },
  }
\tikzset{
	state/.style={
		rectangle,
		draw=black, very thick,
		minimum height=1.0em,
		text centered,
	},
}
\tikzset{
  on each segment/.style={
    decorate,
    decoration={
      show path construction,
      moveto code={},
      lineto code={
        \path [#1]
        (\tikzinputsegmentfirst) -- (\tikzinputsegmentlast);
      },
      curveto code={
        \path [#1] (\tikzinputsegmentfirst)
        .. controls
        (\tikzinputsegmentsupporta) and (\tikzinputsegmentsupportb)
        ..
        (\tikzinputsegmentlast);
      },
      closepath code={
        \path [#1]
        (\tikzinputsegmentfirst) -- (\tikzinputsegmentlast);
      },
    },
  },
  mid arrow/.style={postaction={decorate,decoration={
        markings,
        mark=at position .5 with {\arrow[#1]{stealth}}
      }}},
}

\usepackage{etoolbox}
\makeatletter%
\AfterPreamble{%
	\usepackage{hyperref}%
	\let\oldhypertarget\hypertarget%
	\renewcommand{\hypertarget}[2]{%
		\oldhypertarget{#1}{#2}%
		\protected@write\@mainaux{}{%
			\string\expandafter\string\gdef%
			\string\csname\string\detokenize{#1}\string\endcsname{#2}%
		}%
	}%
	\newcommand{\myhyperlink}[1]{%
		\hyperlink{#1}{\csname #1\endcsname}%
	}%
}
\makeatother%

\newcounter{Remark}
\newcounter{Problem}
\makeatletter
\def\blfootnote{\xdef\@thefnmark{}\@footnotetext}
\makeatother

\newcounter{FootNoteCounter}
\newcommand\copyrighttext{%
	\small \begin{center} \color{red} \vspace{-2em} \textcopyright\,2022 IEEE. Personal use of 
	this material is permitted. Permission from IEEE must be obtained for all other uses. 
	\end{center}}

\title{\copyrighttext \vspace{-0.75em} \LARGE \bf
	A Nonlinear Model Predictive Control Strategy for\\Autonomous Racing of Scale Vehicles
}

\author{Vittorio Cataffo$^1$, Giuseppe Silano$^{2}$, Luigi Iannelli$^1$, Vicenç Puig$^3$, and Luigi Glielmo$^1$
	\thanks{$^1$V.~Cataffo, L.~Iannelli, and L.~Glielmo are with the Department of Engineering, University of Sannio in Benevento, Benevento, Italy (email: {\tt\small 
	\{v.cataffo1, luigi.iannelli, glielmo\}@unisannio.it}).}
	\thanks{$^2$G.~Silano is with Ricerca sul Sistema Enegertico (RSE) S.p.A., Department of Generation Technologies and Materials, Milan, Italy, and also with the Faculty of Electrical Engineering, Czech Technical University in Prague, Prague, Czech Republic (email: {\tt\small giuseppe.silano@fel.cvut.cz}).}
	\thanks{$^3$V.~Puig is with the Center of Supervision, Security and Automatic Control, Universitat Politècnica de Catalunya, Terrassa, Spain (email: {\tt\small vicenc.puig@upc.edu}).}
	\thanks{This work was partially funded by the ECSEL Joint Undertaking (JU) research and innovation programme COMP4DRONES under grant agreement no. 826610, and by the European Union's Horizon 2020 research and innovation programme AERIAL-CORE under grant agreement no. 871479.}
}

\begin{document}

\maketitle
\thispagestyle{empty}
\pagestyle{empty}


\begin{acronym}
    \acro{AV}[AV]{Autonomous Vehicle}
    \acro{CoG}[CoG]{Center of Gravity}
    \acro{DC}[DC]{Direct Current}
    \acro{IoT}[IoT]{Internet of Things}
	\acro{MPC}[MPC]{Model Predictive Control}
	\acro{NMPC}[NMPC]{Nonlinear Model Predictive Control}
	\acro{NLP}[NLP]{Nonlinear Programming}
	\acro{PWM}[PWM]{Pulse Width Modulation}
	\acro{QP}[QP]{Quadratic Programming}
	\acro{wrt}[w.r.t.]{with respect to}
\end{acronym}
	


\begin{abstract}

A~\ac{NMPC} strategy aimed at controlling a small-scale car model for autonomous racing competitions is presented in this paper. The proposed control strategy is concerned with minimizing the lap time while keeping the vehicle within track boundaries. The optimization problem considers both the vehicle's actuation limits and the lateral and longitudinal forces acting on the car modeled through the Pacejka's magic formula and a simple drivetrain model. Furthermore, the approach allows to safely race on a track populated by static obstacles generating collision-free trajectories and tracking them while enhancing the lap timing performance. Gazebo simulations using the F1/10 simulator showcase the feasibility and validity of the proposed control strategy. The code is released as open-source making it possible to replicate the obtained results.

\end{abstract}






\section{Introduction}
\label{sec:introduction}

Over the last years, the need to provide more affordable mobility and to reduce greenhouse gases from needless idling is creating a high expectation environment as a perfect enabler for~\acp{AV} and applications of autonomous driving~\cite{Omeiza2021IEEETITS}.


In an attempt to push the limits towards the development of new technologies, numerous  competitions are organized and held in major international conferences. Above all, the F1/10 Autonomous Racing competition~\cite{Agnihotri2020ACM, Babu2020F1tenth} is one of the most popular; its name derives from the use of 1:10 scaled-down car models. 
Depending on the task objective, the problem is faced with different levels of detail and approximations~\cite{Kong2015IVS, Alcala2020RAS}. 
Several approaches have been proposed in the various editions of the competition~\cite{Alcala2020CEP, Klapalek2021IROS, Verschueren2016ECC}. However, when vehicle nonlinearities are excited, the control task is inevitably more demanding and this opens up new challenges to be solved.
%

The~\ac{MPC} approach has been proved to be a promising solution to control the car motion while complying with its dynamics and multiple heterogeneous constraints~\cite{Rosolia2017IEEETCST, Rosolia2020TCST}. Specifically,~\acf{NMPC} has resulted particularly suitable to control autonomous racing cars when their agility is essential for the particular application and must be exploited at the best~\cite{Guo2018TII, Guoying2022CEP}. However, the intrinsic capability of the framework to embed physical constraints comes with a cost: the high-computational load required to solve the optimization problem. 

Advances in the computational capabilities of modern computers and improvements in the algorithms efficiency~\cite{Verschueren2021, Andersoon2019} have made it possible to manage such complexity along with the real-time requirements to solve these problems. Several software frameworks~\cite{Sathya2018ECC, Chen2019ECC} have been released over the years to facilitate modeling, control design, and simulation for a broad class of~\ac{NMPC} applications. 


Various works have investigated~\ac{NMPC} strategies both as a trajectory generator~\cite{Frasch2013ECC, Alcala2020RAS} and as a tracking controller~\cite{Verschueren2016ECC, Alcala2020CEP}. A common problem is the tracking of the lane center line while avoiding collisions with obstacles placed along the track. In most cases,~\ac{NMPC} is used in the outer loop of a cascaded architecture to provide a reference trajectory to an inner loop tracking controller~\cite{Frasch2013ECC, Alcala2020CEP}. This approach allows to attain the tracking-lane objectives, but it can cause problems since the~\ac{NMPC} generator does not consider the limitations posed by the low-level controller~\cite{Rosolia2020TCST}. As a consequence, the generated trajectory could violate the vehicle's actuator limits resulting in an unfeasible solution.

To overcome this limitation,~\ac{NMPC} can be used to combine trajectory generation, subject to obstacle avoidance constraints, and trajectory tracking, subject to actuation limits and track boundaries, in a single optimization problem~\cite{Rosolia2017IEEETCST, Guo2018TII, Guoying2022CEP}. The so-formulated problem allows to keep tracking of the lane center line, preventing critical configurations that could move the vehicle out of the limited-width track, while taking the actuator limitations into account. 

Following this line of research, an~\ac{NMPC} architecture for lane center line tracking for autonomous racing car competitions is here proposed by considering for the first time, to the best of authors knowledge, the case of 1:10 scale racing cars. The optimization problem considers both vehicle dynamics constraints and physical actuation limits. The car dynamics are described by a bicycle model and longitudinal and lateral forces acting on the vehicle are modeled by a drivetrain model and the Pacejka's magic formula, respectively. An identification problem is set up to obtain the model parameters using experimental data collected in the F1/10 simulator~\cite{Babu2020F1tenth}. The~\ac{NMPC} approach is coded using the OpEn framework and PANOC as solver~\cite{Sathya2018ECC, Pantelis2020IFAC}. Compared to the mentioned approaches, both the optimal racing trajectory and tracking lane problems are solved within the same optimization framework while avoiding static obstacles placed along the track. The code is released as open-source\footnote{\url{https://bit.ly/3vPlmFf}} making it possible to go through any part of the framework and to replicate the obtained results. Illustrative videos with the achieved Gazebo simulations are available at~\url{https://youtu.be/w5c328rQmX4}.




\section{System Modeling}
\label{sec:systemModeling}




\subsection{Vehicle dynamics}
\label{sec:vehicleDynamics}

The prediction model is a key part in~\ac{MPC} laws. One of the most common approaches used in vehicle dynamics applications is to simplify the vehicle model to that of a 2-wheeled bicycle model. This approximation is sufficient to provide the necessary inputs to actuators due to the small size of the vehicle. Similar approaches have been adopted by~\cite{Alcala2020CEP, Liniger2014OPCAM, Verschueren2016ECC, Guoying2022CEP} with different levels of detail and approximation. 

%
\begin{figure}[t]
    \centering
    \scalebox{0.9}{
    \begin{tikzpicture}
        \node[inner sep=0pt] (model) at (0, 0) {\adjincludegraphics[width=0.45\textwidth, keepaspectratio, trim={{.0\width} {.0\height} {.0\width} {.18\height}}, clip]{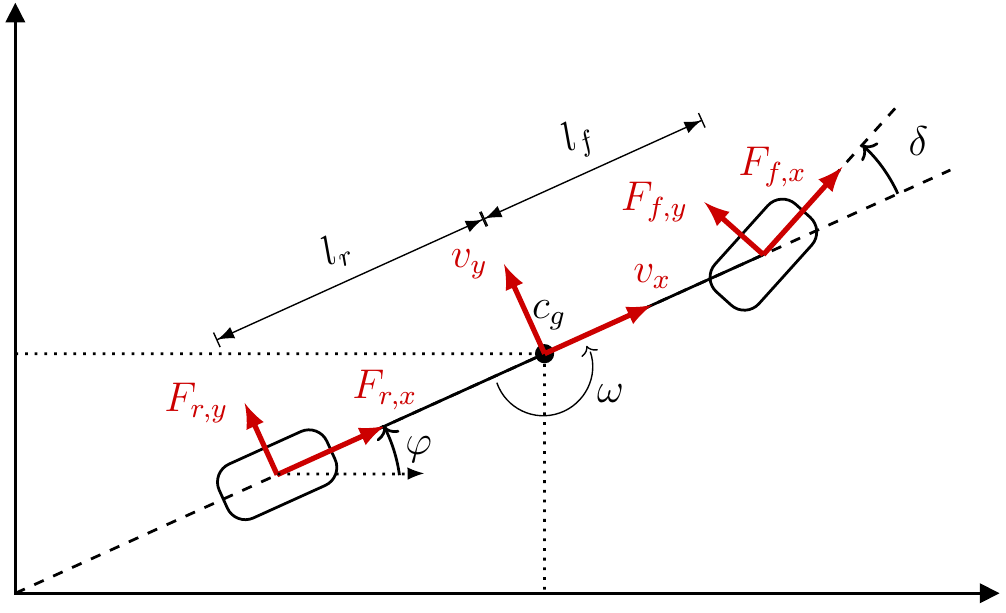}};
        \draw (-3.875,1.9) node[text centered]{$\blacktriangle$};
        \draw (0.45,-2.1) node[text centered]{$p_x$};
        \draw (-4.10,0) node[text centered]{$p_y$};
        \draw (-4.10,-2.0) node[text centered]{$O$};
        \draw (3.85,-2.1) node[text centered]{$X$};
        \draw (-4.10,1.9) node[text centered]{$Y$};
        \draw[white, line width=1.75pt] (-3.8,-1.80) -- (-2.25,-1.10);
        \draw[white, line width=1.75pt] (-2.22,-1.075) -- (-1.84,-0.925);
        \draw[white, line width=1.75pt] (-3.85,-1.795) -- (-3.8,-1.795);
    \end{tikzpicture}
    }
    \vspace{-0.1cm}
    \caption{A representation of the dynamic bicycle model.}
    \label{fig:vehicle_model}
\end{figure}
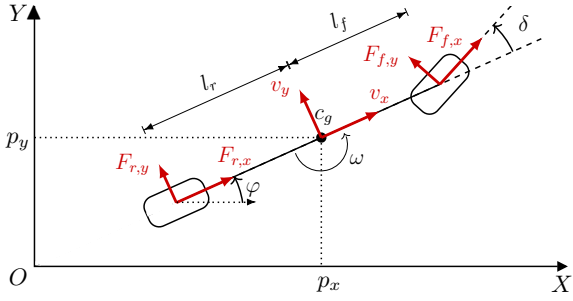
Let us consider the bicycle model as described in~\cite{Kong2015IVS, Liniger2014OPCAM} and depicted in Fig.~\ref{fig:vehicle_model}. The lateral and longitudinal forces $F_{r,y} \in \mathbb{R}$, $F_{r,x} \in \mathbb{R}$, $F_{f,y} \in \mathbb{R}$ and $F_{f,x} \in \mathbb{R}$ describe the forces acting on the tires, where the subscripts $r$ and $f$ refer to the rear and front parts of the vehicle, respectively, while the subscripts $x$ and $y$ denote the longitudinal ($x$) and lateral ($y$) axes along with the forces are exerted. The parameters $l_r \in \mathbb{R}_{\geq 0}$ and $l_f \in \mathbb{R}_{\geq 0}$ represent the distance between the rear and front wheel to the~\ac{CoG} $c_g$, respectively. The steering angle $\delta \in \mathbb{R}$ quantifies the deflection of the front wheel, while the rear wheel is fixed. The vehicle's position and orientation in the \textit{world frame} $\mathcal{F}_W$ are given by $p_x \in \mathbb{R}$ , $p_y \in \mathbb{R}$ and $\varphi \in \mathbb{R}$, respectively. 

The lateral forces $F_{f,y}$ and $F_{r,y}$ acting on the vehicle are described using the simplified Pacejka's magic formula~\cite{Kong2015IVS, Liniger2014OPCAM} as
\begin{subequations}\label{eq:magic_formula}
    \begin{align}
    F_{f,y} &= D_f \sin \left( {C_f \arctan \left( B_f \alpha_f \right)} \right), \\
    F_{r,y} &= D_r \sin \left( {C_r \arctan \left(B_r \alpha_r \right)} \right), 
    \end{align}
\end{subequations}
where $B_f \in \mathbb{R}$ and $B_r \in \mathbb{R}$ are the \textit{stiffness factors}, $C_f \in \mathbb{R}$ and $C_r \in \mathbb{R}$ are the \textit{shape factors}, and $D_f \in \mathbb{R}$ and $D_r \in \mathbb{R}$ are the \textit{peak factors}. Such an approximation allows meeting the trade‐off between precision and computational requirements. The slip angles $\alpha_f \in \mathbb{R}$ and $\alpha_r \in \mathbb{R}$ are described as
\begin{subequations}\label{eq:slipAngles}
    \begin{align}
    \alpha_{f} &= -\arctan \left(\frac{ \omega l_{f} + v_{y}}{v_{x}} \right) + \delta, \\
    \alpha_{r} &= \arctan \left(\frac{ \omega l_{r} - v_{y}}{v_{x}} \right),
    \end{align}
\end{subequations}
where $\omega \in \mathbb{R}$ represents the change rate of the orientation $\varphi$ during time. Equations~\eqref{eq:magic_formula} and~\eqref{eq:slipAngles} model the interaction between the car and the road. 

%

For ease of modeling, the longitudinal forces $F_{f,x}$ and $F_{r,x}$ are assumed to be equal, i.e., $F_{f,x} = F_{r,x} = F_x$, and described using the following \textit{drivetrain model}~\cite{Liniger2014OPCAM},
\begin{equation}\label{eq:driveTrainModel}
    F_x = (C_{m1} - C_{m2}v_x)\tilde{d} - C_{m3} - C_{m4}v_x^2,
\end{equation}
where $\tilde{d} \in [0,1]$ is the~\ac{PWM} signal applied to motors and $C_{m1} \in \mathbb{R}_{\geq 0}$, $C_{m2} \in \mathbb{R}_{\geq 0}$, $C_{m3} \in \mathbb{R}_{\geq 0}$ and $C_{m4} \in \mathbb{R}_{\geq 0}$ are empirical parameters used to shape the model's response curve to fit the drivetrain characteristics~\cite{Liniger2014OPCAM}. The driving command $\tilde{d}=1$ corresponds to full throttle, while $\tilde{d}=0$ to full braking.


Hence, using Newton's second law, the vehicle dynamics~\ac{wrt} $c_g$ can be described as
\begin{equation}\label{eq:bicycleModelEquations}
\left\{
	\begin{array}{l}
    \dot{p}_x = v_x \cos{\varphi} - v_y \sin{\varphi} \\
    \dot{p}_y = v_x \sin{\varphi} + v_y \cos{\varphi} \\
    \dot{\varphi} = \omega \\
    m\dot{v}_x =  F_{r,x} - F_{f,y} \sin{\delta} + F_{f,x} \cos{\delta} + m v_y \omega \\
    m\dot{v}_y = F_{r,y} + F_{f,y} \cos{\delta} + F_{f,x} \sin{\delta} - m v_x \omega \\
    J_z \dot{\omega} = l_f F_{f,y} \cos{\delta} + l_f F_{f,x} \sin{\delta} - l_r F_{r,y}
	\end{array}
\right.,
\end{equation}
where $m \in \mathbb{R}_{> 0}$ is the mass of the vehicle, $J_z \in \mathbb{R}_{>0}$ is the $z$-component of the inertia matrix $\mathbf{J} \in \mathbb{R}_{\geq 0}^{3 \times 3}$, and $v_x$ and $v_y$ represent the car's velocity along the $x$-axis and $y$-axis of the body frame $\mathcal{F}_B$, respectively. Note that, with abuse of notation, the longitudinal forces $F_{f,x}$ and $F_{r,x}$~\eqref{eq:driveTrainModel} are left in the model description~\eqref{eq:bicycleModelEquations} as the simplification $F_{f,x}=F_{r,x}=F_x$ does not affect the system modeling.

The model~\eqref{eq:bicycleModelEquations} describes a nonlinear dynamic system $\dot{\mathbf{x}} = f_c(\mathbf{x}, \mathbf{u})$, with state $\mathbf{x} = [p_x, p_y, \varphi, v_x, v_y, \omega]^\top \in \mathbb{R}^6$ and control input $\mathbf{u} = [\tilde{d}, \delta]^\top \in \mathbb{R}^2$. 





\subsection{System identification}
\label{sec:systemIdentification}

In order to exploit the prediction model, it is pivotal to identify the model's parameters, i.e., those describing the lateral and longitudinal forces modeled by the simplified Pacejka's magic formula~\eqref{eq:magic_formula} and the drivetrain model~\eqref{eq:driveTrainModel}. The set of parameters to be identified can be indicated as the vector $\bm{\zeta}  \in \mathbb{R}^{10}$, and specifically $\bm{\zeta} = [B_f, B_r, C_f, C_r, D_f, D_r, C_{m1}, C_{m2}, C_{m3}, C_{m4}]^\top$. As for the remaining model's parameters, they are assumed to be known for the particular vehicle. Table~\ref{tab:modelParameterValues} reports the list of parameters along with their values.

Acceleration and deceleration experiments in the F1/10 simulator can be used for the identification process based on a least squares minimization approach.


By assuming having $M \in \mathbb{N}_{>0}$  signal samples acquired at a fixed sampling rate, the sequences of the linear velocities $v_x$ and $v_y$ and the angular velocity $\omega$, can be denoted, respectively, by the vectors $\mathbf{v}_x= [v_{x_1}, v_{x_2}, \cdots, v_{x_M}]^\top $, $\mathbf{v}_y = [v_{y_1}, v_{y_2}, \cdots, v_{y_M}]^\top $, and $\bm{\omega} = [\omega_1, \omega_2, \cdots, \omega_M]^\top $.  Thus, the least-squares minimization problem becomes 
%
    \begin{align}
    &\hspace{-0.27cm}\min_{\underline{\bm{\zeta}} \leq \bm{\zeta} \leq \bar{\bm{\zeta}}
    }
    { \lVert \mathbf{v}_x - \hat{\mathbf{v}}_x(\bm{\zeta}) \rVert^2 + \lVert \mathbf{v}_y - \hat{\mathbf{v}}_y(\bm{\zeta}) \lVert^2 + \lVert \bm{\omega} - \hat{\bm{\omega}}(\bm{\zeta}) \rVert^2 }, \label{eq:leastSquareProblem} 
    %
    %
    \end{align}
%
with $\hat{\mathbf{v}}_x(\bm{\zeta}) \in \mathbb{R}^{M}$, $\hat{\mathbf{v}}_y(\bm{\zeta}) \in \mathbb{R}^{M}$ and $\hat{\bm{\omega}}(\bm{\zeta}) \in \mathbb{R}^{M}$ representing the one-step prediction linear and angular velocities based on the discretized vehicle's model~\eqref{eq:bicycleModelEquations}, while $\underline{\bm{\zeta}}$ and $\bar{\bm{\zeta}}$ denoting the range of minimum and maximum admissible values, respectively, for the parameters $\bm{\zeta}$.
\begin{table}[tb]
    \scalebox{0.85}{
    \centering
    \begin{tabular}{|l|l|l|l|l|l|}
    \hline
    \textbf{Sym.} & \textbf{Value} & \textbf{Sym.} & \textbf{Value} & \textbf{Sym.} & \textbf{Value} \\

    \hline \hline
    $l_f$ & $\SI{0.178}{\meter}$ & $l_r$ & $\SI{0.147}{\meter}$ & $m$ & $\SI{5.692}{\kilogram}$\\
    $J_z$ & $\SI{0.204}{\kilogram\meter\squared}$ & $B_f$ & $\num{9.242}$ & $B_r$ & $\num{17.716}$ \\
    $C_f$ & $\num{0.085}$ & $C_r$ & $\num{0.133}$ & $D_f$ & $\SI{134.585}{\newton}$ \\
    $D_r$ & $\SI{159.919}{\newton}$ & $C_{m1}$ & $\SI{20}{\newton}$ & $C_{m2}$ & $\SI{6.92e-7}{\kilogram\per\second}$ \\
    $C_{m3}$ & $\SI{3.99}{\newton}$ & $C_{m4}$ & $\SI{0.67}{\kilogram\per\meter}$ & $M$ & $\num{1674}$ \\
    \hline
    \end{tabular}
    }
    \caption{Model's parameter values obtained through the identification procedure with signals acquired every $\SI{50}{\milli\second}$.}
    \label{tab:modelParameterValues}
\end{table}

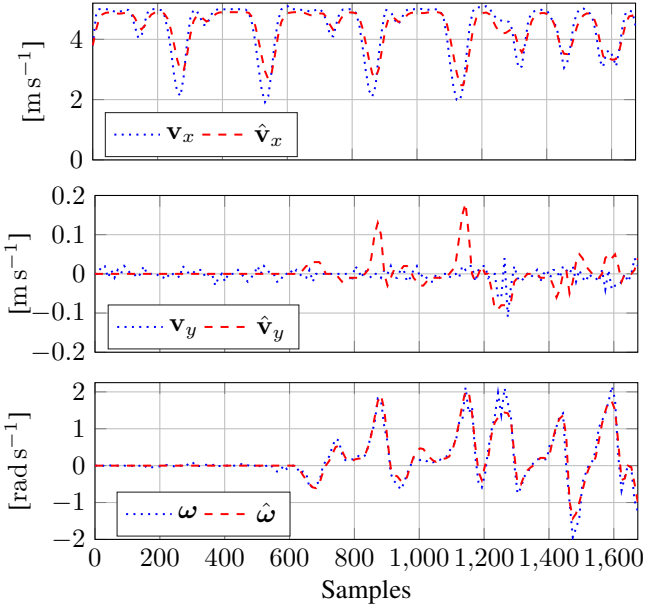
\begin{figure}[tb]
    \hspace{-0.05cm}
    \begin{subfigure}{0.95\columnwidth}
        \centering
        \begin{tikzpicture}
    	\begin{axis}[%
    	width=2.8119in,%
    	height=0.8183in,%
    	at={(0.758in,0.481in)},%
    	scale only axis,%
    	xmin=0,%
    	xmax=1674,%
    	ymax=5.200,%
    	ymin=0,%
    	xmajorgrids,%
    	ymajorgrids,%
    	xticklabels={,,},%
    	ylabel style={yshift=-0.4215cm}, 
    	xlabel style={yshift=0.115cm}, 
    	ylabel={[$\si{\meter\per\second}$]},%
    	axis background/.style={fill=white},%
    	legend style={at={(0.2,0.305)},anchor=north,legend cell 
    		align=left,draw=none,legend columns=-1,align=left,draw=white!15!black}
    	]
    	\addplot [color=blue, dotted, line width=0.75pt]
    	   file{matlabPlots/identification/v_x_measured.txt};%
    	\addplot [color=red, dashed, line width=0.75pt]                                 file{matlabPlots/identification/v_x_identified.txt};%
    	\legend{$\mathbf{v}_x$, $\hat{\mathbf{v}}_x$};%
    	\end{axis}
    	\end{tikzpicture}
    \end{subfigure}
    \\
    \vspace{0.05cm}
    \hspace{-0.35cm}
    \begin{subfigure}{0.95\columnwidth}
        \centering
        \begin{tikzpicture}
    	\begin{axis}[%
    	width=2.8119in,%
    	height=0.8183in,%
    	at={(0.758in,0.481in)},%
    	scale only axis,%
    	xmin=0,%
    	xmax=1674,%
    	ymax=0.2,%
    	ymin=-0.2,%
    	xmajorgrids,%
    	ymajorgrids,%
    	xticklabels={,,},%
    	ylabel style={yshift=-0.215cm}, 
    	xlabel style={yshift=0.115cm}, 
    	ylabel={[$\si{\meter\per\second}$]},%
    	axis background/.style={fill=white},%
    	legend style={at={(0.2,0.305)},anchor=north,legend cell 
    		align=left,draw=none,legend columns=-1,align=left,draw=white!15!black}
    	]
    	\addplot [color=blue, dotted, line width=0.75pt]
    	   file{matlabPlots/identification/v_y_measured.txt};%
    	\addplot [color=red, dashed, line width=0.75pt]                                 file{matlabPlots/identification/v_y_identified.txt};%
    	\legend{$\mathbf{v}_y$, $\hat{\mathbf{v}}_y$};%
    	\end{axis}
    	\end{tikzpicture}
    \end{subfigure}
    \\
    \vspace{0.05cm}
    \hspace{-0.35cm}
    \begin{subfigure}{0.95\columnwidth}
        \centering
        \begin{tikzpicture}
    	\begin{axis}[%
    	width=2.8119in,%
    	height=0.8183in,%
    	at={(0.758in,0.481in)},%
    	scale only axis,%
    	xmin=0,%
    	xmax=1674,%
    	ymax=2.25,%
    	ymin=-2.0,%
    	xmajorgrids,%
    	ymajorgrids,%
    	ylabel style={yshift=-0.215cm}, 
    	xlabel style={yshift=0.115cm}, 
    	xlabel={Samples},%
    	ylabel={[$\si{\radian\per\second}$]},%
    	axis background/.style={fill=white},%
    	legend style={at={(0.2,0.305)},anchor=north,legend cell 
    		align=left,draw=none,legend columns=-1,align=left,draw=white!15!black}
    	]
    	\addplot [color=blue, dotted, line width=0.75pt]
    	   file{matlabPlots/identification/omega_measured.txt};%
    	\addplot [color=red, dashed, line width=0.75pt]                                 file{matlabPlots/identification/omega_identified.txt};%
    	\legend{$\bm{\omega}$, $\hat{\bm{\omega}}$};%
    	\end{axis}
    	\end{tikzpicture}
    \end{subfigure}
    \vspace{-1em}
    \caption{A comparison between the identified (hat symbols) and collected linear and angular velocity values.}
    \label{fig:comparisonMeasuredIdentified}
\end{figure}

The minimization problem~\eqref{eq:leastSquareProblem} was coded using the 2019b release of MATLAB and solved using the \texttt{fmincon} function of the MathWorks Optimization Toolbox. Figure~\ref{fig:comparisonMeasuredIdentified} shows the comparison between the acquired ($\mathbf{v}_x$, $\mathbf{v}_y$ and $\bm{\omega}$) and the predicted ($\hat{\mathbf{v}}_x$, $\hat{\mathbf{v}}_y$ and $\hat{\bm{\omega}}$) linear and angular velocities corresponding to the identified values of parameters.

\section{Problem Formulation}
\label{sec:problemFormulation}

A small-scale racing car is required to race along a track minimizing the lap time and remaining within the track boundaries. Meanwhile, the car is required to safely compete avoiding static obstacles populating the track. In addition, the control system is demanded to comply with the vehicle's actuation limits while fulfilling the mission objectives.

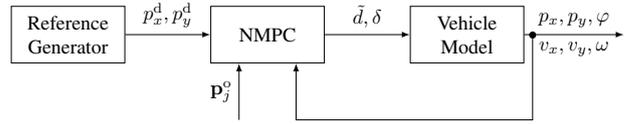
\begin{figure}[tb]
    \centering
    \centering
	\scalebox{0.75}{
		\begin{tikzpicture}
		
		\node (ReferenceGenerator) at (-3.5,0) [draw, rectangle, text centered, minimum width=1.5cm, minimum height=1cm, text width=5em]{Reference\\Generator};
		
		\node (NMPC) at (0,0) [draw, rectangle, text centered, minimum width=1cm, minimum height=1cm, text width=5em]{NMPC};
		
		\node (Car) at (3.5,0) [draw, rectangle, text centered, minimum width=1cm, minimum height=1cm, text width=5em]{Vehicle\\Model};
		
		\draw[-latex] (ReferenceGenerator) -- node[above]{$p_x^\mathrm{d}, p_y^\mathrm{d}$} (NMPC);
		\draw[-latex] (NMPC) -- node[above]{$\tilde{d}, \delta$} (Car);
		\draw[-latex] (Car) -- node[above]{$p_x, p_y, \varphi$} node[below]{$v_x, v_y, \omega$} ($(Car) + (2.75, 0) $);
		\draw[-latex] ($(NMPC.south) - (0.5, 1.0)$) -- node[left]{$\mathbf{p}^\mathrm{o}_j$} ( $(NMPC.south) - (0.5, 0)$ );
		\draw[-latex] ($(Car.south) + (1.15,0.5) $) -- ( $(Car.south) - (-1.15,1)$ ) -- ( $(Car.south) - (3.0,1)$ ) -- ( $(Car.south) - (3.0,0)$ ); 
		\node at  ($(Car.south) + (1.15,0.5) $) [circle,fill,draw,inner sep=0pt,minimum size=3pt, text centered]{};
		
		\end{tikzpicture}
	}
    \caption{Block diagram of the proposed control strategy.}
    \label{fig:controlArchitecture}
\end{figure}

A two-layer control architecture has been proposed to cope with the problem of making a small-scale car race along a track minimizing the lap time and remaining within the track boundaries, by avoiding static obstacles, as well. A reference generator algorithm provides the reference point coordinates ($p_x^\mathrm{d}$, $p_y^\mathrm{d}$) to an~\ac{NMPC} tracking controller which computes the control signals ($\tilde{d}$, $\delta$) to reach the target point while satisfying all constraints. Figure~\ref{fig:controlArchitecture} describes the overall control system architecture.

The reference generator supplies reference point coordinates ($p_x^\mathrm{d}$, $p_y^\mathrm{d}$) to the~\ac{NMPC} tracking controller at each instance of the optimal problem. The planner leverages the capability of computing the projection of the car's position along the lane center line. For ease of experimentation, the proposed solution assumes to know the shape of the track, and therefore the lane center line. Such an assumption is in line with the competition rules~\cite{Agnihotri2020ACM} on which the algorithm was designed.

Let us assume the lane center line is sampled in $K \in \mathbb{N}_{>0}$ points with a constant sampling space (Euclidean distance) $d_s \in \mathbb{R}_{>0}$ and denote with $\mathbf{p}^\mathrm{c}_k = [p_{x_k}^\mathrm{c}, p_{y_k}^\mathrm{c}]^\top \in \mathbb{R}^2$ the $k$-th element of the lane center line, with $k\in\{1,2, \ldots, K\}$. Hence, the lane center line can be represented as the sequence $\mathbf{p}^\mathrm{c} = \{\mathbf{p}_1^\mathrm{c}, \mathbf{p}_2^\mathrm{c}, \ldots, \mathbf{p}_K^\mathrm{c} \}$. Let us also define with $\mathbf{p} = [p_x, p_y]^\top \in \mathbb{R}^2$ the current car's position along the track. At each control step, the reference generator projects the car's position onto the center line:
\begin{equation}\label{eq:argminPlanner}
   i^*= 
    \argmin_{i} \lVert \mathbf{p}^\mathrm{c}_i - \mathbf{p} \lVert^2, \quad  \begin{bmatrix}
    p_x^\prime \\
    p_y^\prime
    \end{bmatrix} = \mathbf{p}_{i^*}^\mathrm{c}.
\end{equation}
%

\begin{figure}[tb]
    \centering
    \begin{subfigure}{0.45\columnwidth}
       \begin{tikzpicture}
        \node[inner sep=0pt] (model) at (0, 0) {\adjincludegraphics[width=\textwidth, keepaspectratio, trim={{.0\width} {.0\height} {.0\width} {.0\height}}, clip]{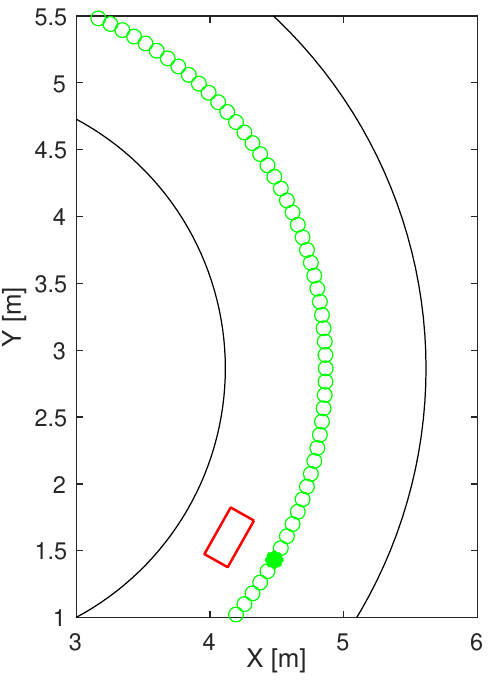}};
        \draw (0.85,-1.7) node[text centered]{\footnotesize{$(p_x^\prime, p_y^\prime)$}};
        \draw (-0.8,-1.4) node[text centered]{\footnotesize{$(p_x, p_y)$}};
        \node at (-0.15,-1.5) [circle,fill,draw,red,inner sep=0pt,minimum size=3pt, text centered]{}; 
        \draw[dashed, line width=0.75] (-0.15,-1.5) -- (0.2,-1.7);
       \end{tikzpicture} 
    \end{subfigure}
    \hspace{0.05cm}
    \begin{subfigure}{0.45\columnwidth}
        \begin{tikzpicture}
        \node[inner sep=0pt] (model) at (0, 0) {\adjincludegraphics[width=\textwidth, keepaspectratio, trim={{.0\width} {.0\height} {.0\width} {.0\height}}, clip]{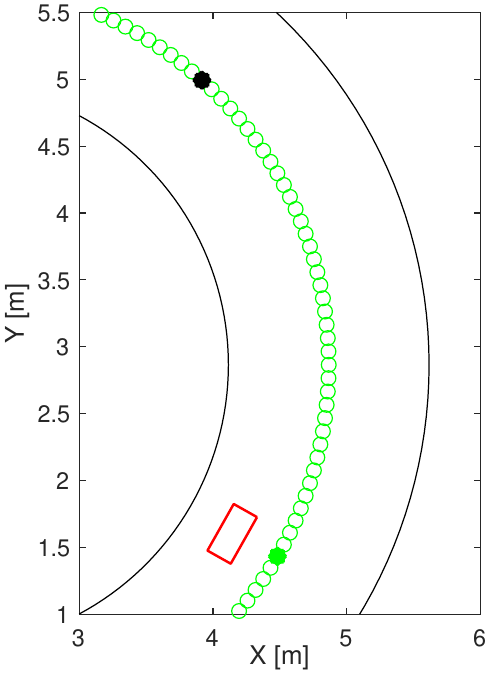}};
        \draw (0.85,-1.7) node[text centered]{\footnotesize{$(p_x^\prime, p_y^\prime)$}};
        \draw (0.35,2.00) node[text centered]{\footnotesize{$(p_x^\mathrm{d}, p_y^\mathrm{d})$}};
        \draw (-0.8,-1.4) node[text centered]{\footnotesize{$(p_x, p_y)$}};
        \node at  (-0.15,-1.5) [circle,fill,draw,red,inner sep=0pt,minimum size=3pt, text centered]{};
       \end{tikzpicture} 
    \end{subfigure}
    \vspace{-0.75em}
    \caption{In the figure, the car's position ($p_x$, $p_y$), the car's projection on the lane center line ($p_x^\prime$, $p_y^\prime)$ and the reference coordinate ($p_x^\mathrm{d}$, $p_y^\mathrm{d}$) $P$ samples looking-ahead~\ac{wrt} the car's center lane projection.}
    \label{fig:pathPlanningInstances}
\end{figure}
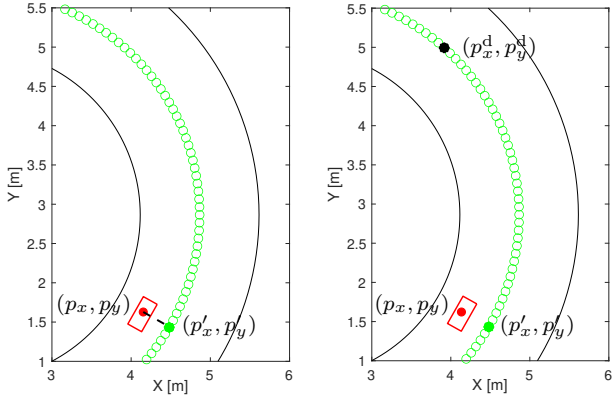

Afterwards, the planner computes the reference coordinates $(p_x^\mathrm{d}$, $p_y^\mathrm{d})$ looking-ahead $P$ waypoints along the lane center line following a variation of the well-known pure-pursuit approach~\cite{Atoui2021MED}. Figure~\ref{fig:pathPlanningInstances} shows a schematic representation of the overall process for a single instance of the reference generator process.

Note that the projection operation and the reference coordinates computation are affected by the number of samples $K$ and the number of look-ahead waypoints $P$. Those are key parameters that can be tuned to achieve the best performance and push the vehicle towards the limits trying, at the same time, not to increase the computation burden.
Table~\ref{tab:controlParameters} reports the values used for the numerical simulation along with the~\ac{NMPC} controller parameters.
\begin{table}[tb]
    \begin{center}
    \begin{tabular}{|l|l|l|l|l|l|}
    \hline
    \textbf{Sym.} & \textbf{Value} & \textbf{Sym.} & \textbf{Value} & \textbf{Sym.} & \textbf{Value} \\
    \hline \hline
    $P$ & $\si{90}$ & $T_s$ & $\SI{0.033}{\second}$ & $N$ & $\num{50}$ \\
    $R_c$ & $\SI{0.24}{\meter}$ & $R_g$ & $\SI{2}{\meter}$ & $\Gamma$ & $\SI{1.5}{\meter}$ \\
    $\mathbf{Q}_1$ & $\text{diag}(10,10)$ & $\mathbf{Q}_2$ & $\text{diag}(10,10)$ &  $R_j$ & $\SI{1}{\meter}$ \\
    $\underline{\bm{\gamma}}_{v_x}$ & 0 & $\bar{\bm{\gamma}}_{v_x}$ & 5 & $d_s$ & $\SI{0.1}{\meter}$\\ 
    $\underline{\bm{\mu}}$ & ($0,-\nicefrac{\pi}{6})^\top$ & $\bar{\bm{\mu}}$ & ($1,\nicefrac{\pi}{6})^\top$ & - & - \\
    \hline
    \end{tabular}
    \caption{Control parameters and minimum ($\underline{\bm{\gamma}}_{v_x}$) and maximum ($\bar{\bm{\gamma}}_{v_x}$) admissible values of the state variable $v_x$.}
    \label{tab:controlParameters}
    \end{center}
\end{table}
%
%
%
%

\begin{figure}[tb]
    \centering
    \begin{tikzpicture}
        \node[inner sep=0pt] (model) at (0, 0) {\adjincludegraphics[width=0.275\textwidth, keepaspectratio, trim={{.0\width} {.0\height} {.0\width} {.0\height}}, clip]{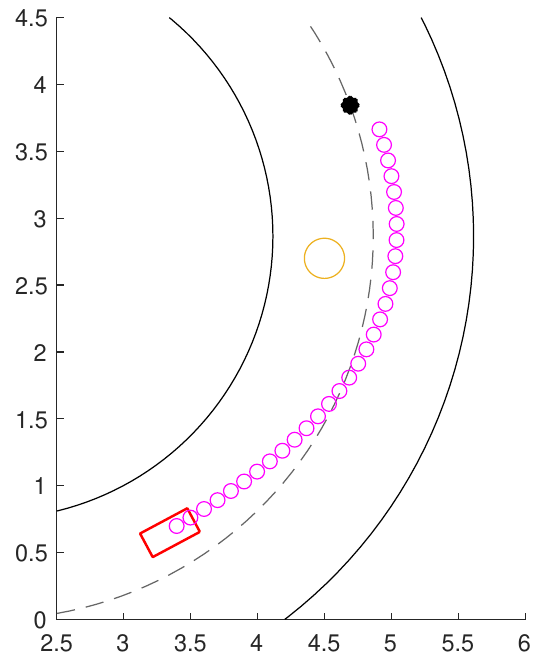}};
        \draw (0.05,2.00) node[text centered]{\footnotesize{$(p_x^\mathrm{d}, p_y^\mathrm{d})$}};
        \draw (-0.9,-1.4) node[text centered]{\footnotesize{$(p_x, p_y)$}};
        \draw (0.3,-3.15) node[text centered]{$\mathrm{X\;[m]}$};
        \draw (-2.65,0.3) node[text centered, rotate=90]{$\mathrm{Y\;[m]}$};
        \node at (0.485,0.675) [circle,fill,draw,brown,inner sep=0pt,minimum size=2pt, text centered]{};
        \draw[densely dotted] (0.485,0.675) -- node[above]{\footnotesize{$R_j$}} (0.705,0.675);
       \end{tikzpicture} 
    \vspace{-1em}
    \caption{The collision-free trajectory (purple) computed by the~\ac{NMPC} tracking controller (a static obstacle in brown). 
    }.
    \label{fig:shapeObstaclesCar}
\end{figure}
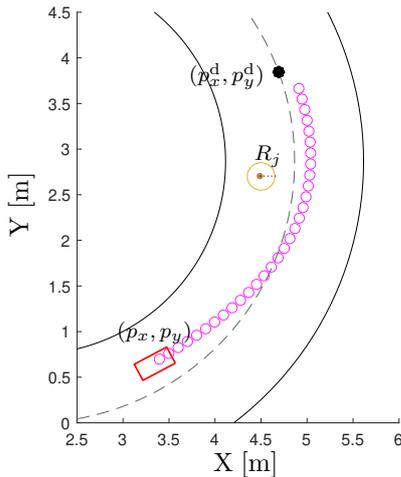

The car is required to minimize the lap time while avoiding static obstacles placed along the track. Positions and size of the obstacles are assumed to be known in the whole prediction horizon of the~\ac{NMPC}. Besides, the shape of the obstacles is approximated with that of a circle of radius $R_j \in \mathbb{R}_{> 0}$, where $j \in \{1, 2, \dots, O\}$, with $O \in \mathbb{N}$ the number of obstacles along the track. A schematic representation is depicted in Fig.~\ref{fig:shapeObstaclesCar}.

The collision avoidance constraint is formulated through the positions of the vehicle $\mathbf{p}$ and the $j$-th obstacle $\mathbf{p}_j^\mathrm{o}$:
\begin{align}\label{eq:obstacleConstraintSquare}
    \lVert \mathbf{p} - \mathbf{p}^\mathrm{o}_j \rVert^2 \geq \Gamma_j^2,
\end{align}
where $\Gamma_j \in \mathbb{R}_{> 0}$ is a threshold distance value that the vehicle has to maintain to avoid collisions with obstacles. The latter is defined accounting for the obstacle sizes and the vehicle's body dimensions in order to ensure a sufficient room margin for maneuvers while remaining within the track boundaries.

Track boundaries constraints are also embedded into the optimal control formulation to maintain the vehicle within the limited-width track. By considering the vehicle's position $\mathbf{p} = [p_x, p_y]^\top$ and its projection on the lane center line $\mathbf{p}^\prime = [p^\prime_x, p^\prime_y]^\top$, the constraint can be formulated as follows
\begin{equation}\label{eq:constraintTrackBoundaries}
    \lVert \mathbf{p} - \mathbf{p}^\prime \rVert^2  \leq ( R_g - R_c )^2,
\end{equation}
where $R_g$ and $R_c$ are defined considering the track width and the car dimensions.





Thus, the optimal control problem, with a prediction horizon of $N \in \mathbb{N}_{>0}$ steps, is described as the minimization of the distance between the last predicted vehicle's position $\mathbf{p}_N$ and the reference point coordinates $\mathbf{p}^\mathrm{d}$. Therefore, at each time step $t_i = i T_s$, with $i \in \mathbb{N}_{>0}$ and $T_s$ being the sampling time, it can be formulated an optimization problem as follows
\begin{subequations}\label{eq:nmpcProblem}
    \begin{align}
    &\hspace{-0.2cm}\minimize_{\mathbf{u} } \;\;
    { \lVert \mathbf{p}_N - \mathbf{p}^\mathrm{d} \rVert^2_{\mathbf{Q}_1} + \sum\limits_{k=0}^{N-1} \;\;  \lVert \mathbf{u}_k - \mathbf{u}_{k-1} \lVert^2_{\mathbf{Q}_2} } \label{subeq:objectiveFunctionNMPC} \\
    %
    &\;\; \text{s.t.}~\; \nonumber \\
    &\hspace{-0.6cm}\;\;\;\;\; \quad \mathbf{u}_{-1} = \mathbf{u}(t_{i-1}),\label{subeq:initialcontrolNMPC}\\
    &\hspace{-0.6cm}\;\;\;\;\;  \quad \mathbf{x}_0 = \mathbf{x}(t_i), \label{subeq:initialstateNMPC} \\
    &\hspace{-0.6cm}\;\;\;\;\; \quad  \mathbf{x}_{k+1} = f(\mathbf{x}_k, \mathbf{u}_k), \ k = 0, \ldots, N-1 , \label{subeq:sysDynamic} \\
    &\hspace{-0.6cm}\;\;\;\;\;  \quad \lVert \mathbf{p}_k - \mathbf{p}^\mathrm{o}_{j_k} \rVert^2 \geq \Gamma_j^2, \ k = 0, \ldots, N, \ j \in \{1,\ldots,  O\}, \label{subeq:obstacleAvoidance} \\
    &\hspace{-0.6cm}\;\;\;\;\;  \quad \lVert \mathbf{p}_k - \mathbf{p}^\prime_k \rVert^2  \leq ( R_g - R_c )^2, \ k =0, \ldots, N - 1 , \label{subeq:trackBoundaries} \\
    &\hspace{-0.6cm}\;\;\;\;\;  \quad \underline{\bm{\mu}} \leq \mathbf{u}_k \leq \bar{\bm{\mu}}, \ k = 0, \ldots, N-1, \label{subeq:uBound} \\
    &\hspace{-0.6cm}\;\;\;\;\;  \quad \underline{\bm{\gamma}} \leq \mathbf{x}_k \leq \bar{\bm{\gamma}}, \ k =0, \ldots, N, \label{subeq:xBound}
    \end{align}
\end{subequations}
where~\eqref{subeq:objectiveFunctionNMPC} is the objective function with $\mathbf{Q}_1, \mathbf{Q}_2  \in \mathbb{R}^{2 \times 2}$ being diagonal weighting matrices, $\mathbf{x}_k$ and $\mathbf{u}_k$ are the sampled predicted state and control input, respectively, at the $k$-th sample of the current MPC interval,~\eqref{subeq:initialcontrolNMPC} and~\eqref{subeq:initialstateNMPC}  initialize the control and the state,~\eqref{subeq:sysDynamic} describes the discretized dynamic model for the vehicle~\eqref{eq:bicycleModelEquations}, and~\eqref{subeq:uBound} and~\eqref{subeq:xBound} are the control input and state limits, respectively. 

The problem~\eqref{eq:nmpcProblem} was encoded using a \textit{single shooting implementation} with the track boundaries~\eqref{eq:constraintTrackBoundaries} and obstacle avoidance~\eqref{eq:obstacleConstraintSquare} constraints treated using the \textit{augmented Lagrangian} and the \textit{penalty method} approaches, respectively, following the constraints formulation of the OpEn framework\footnote{\url{https://alphaville.github.io/optimization-engine}}~\cite{Sathya2018ECC, Pantelis2020IFAC}. The vehicle's dynamics were integrated using a Forward Euler integration method with a sampling time $T_s=\SI{33}{\milli\second}$. The prediction horizon considers $N=50$ steps, which gives an ahead prediction of $\SI{1.65}{\second}$.




\section{Simulation Results}
\label{sec:simulationResults}

To demonstrate the validity of the proposed control strategy, numerical simulations using the F1/10 simulator were performed. The~\ac{NMPC} strategy was coded using the OpEn framework and PANOC as solver~\cite{Sathya2018ECC, Pantelis2020IFAC}. All simulations were performed on a laptop with an i7-10750H processor ($\SI{2.60}{\giga\hertz}$) and $16$GB of RAM running on Ubuntu 18.04 alongside the Melodic Morenia release of ROS. The control algorithm runs at $\SI{30}{\hertz}$ sampling rate. Videos with the simulations are available at~\url{https://youtu.be/w5c328rQmX4}, while the open-source code can be found at~\url{https://bit.ly/3vPlmFf}.

\begin{figure*}[t]
    \vspace{0.5em}
    \hspace{-0.12cm}
    \begin{subfigure}{0.33\columnwidth}
        \centering
        \begin{tikzpicture}
        \node[inner sep=0pt] (model) at (0, 0) {\adjincludegraphics[width=1.9\textwidth, keepaspectratio, trim={{.05\width} {.0\height} {.05\width} {.07\height}}, clip]{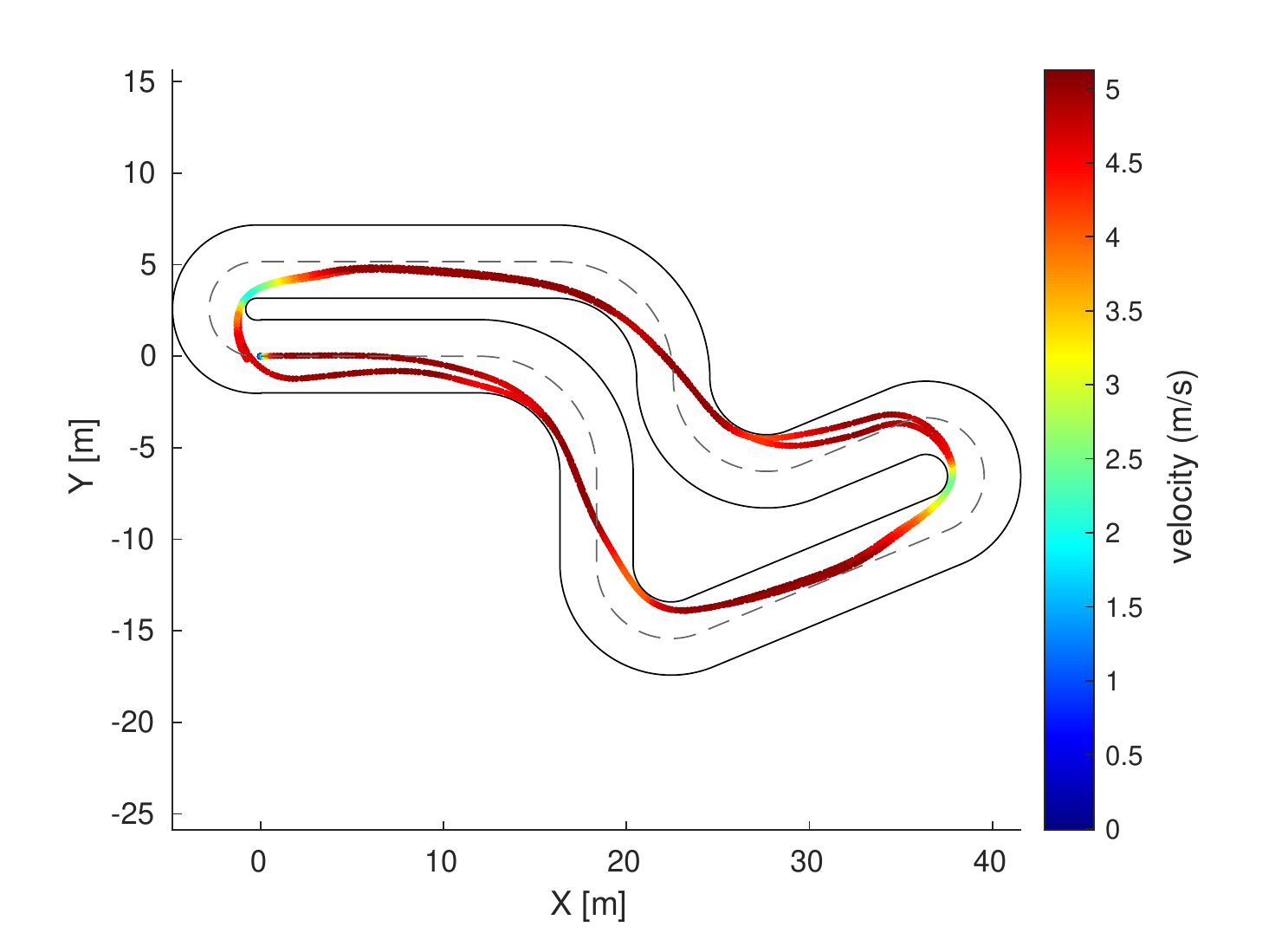}};    
        \end{tikzpicture}
		\label{subfig:senario1NoObstacles}
    \end{subfigure}
    \hspace{2.9cm}
    \begin{subfigure}{0.33\columnwidth}
        \centering
        \begin{tikzpicture}
        \node[inner sep=0pt] (model) at (0, 0) {\adjincludegraphics[width=1.9\textwidth, keepaspectratio, trim={{.05\width} {.0\height} {.05\width} {.07\height}}, clip]{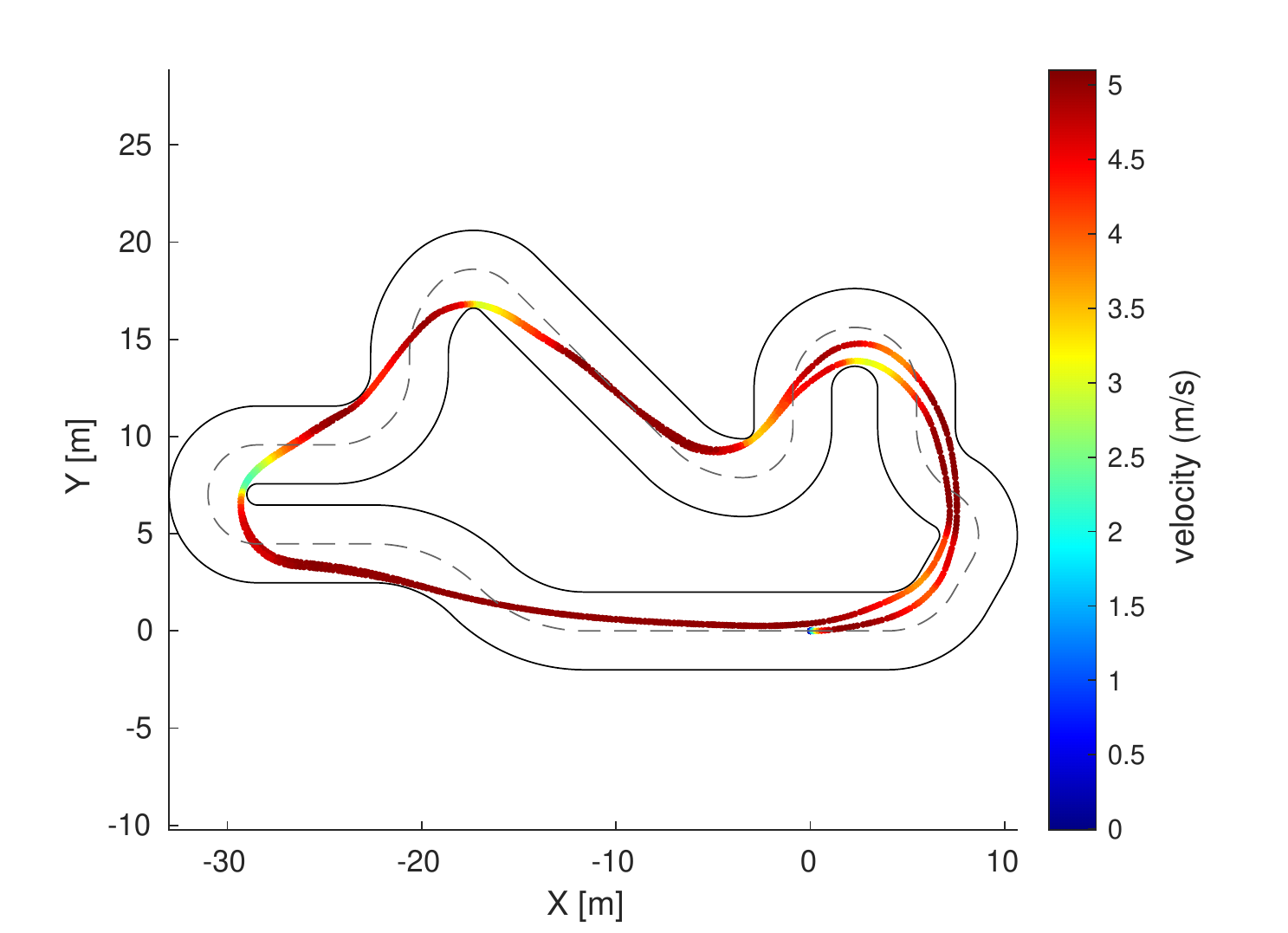}};    
        \end{tikzpicture}
		\label{subfig:senario2NoObstacles}
    \end{subfigure}
    \hspace{2.9cm}
    \begin{subfigure}{0.33\columnwidth}
        \centering
        \begin{tikzpicture}
        \node[inner sep=0pt] (model) at (0, 0) {\adjincludegraphics[width=1.9\textwidth, keepaspectratio, trim={{.05\width} {.0\height} {.05\width} {.07\height}}, clip]{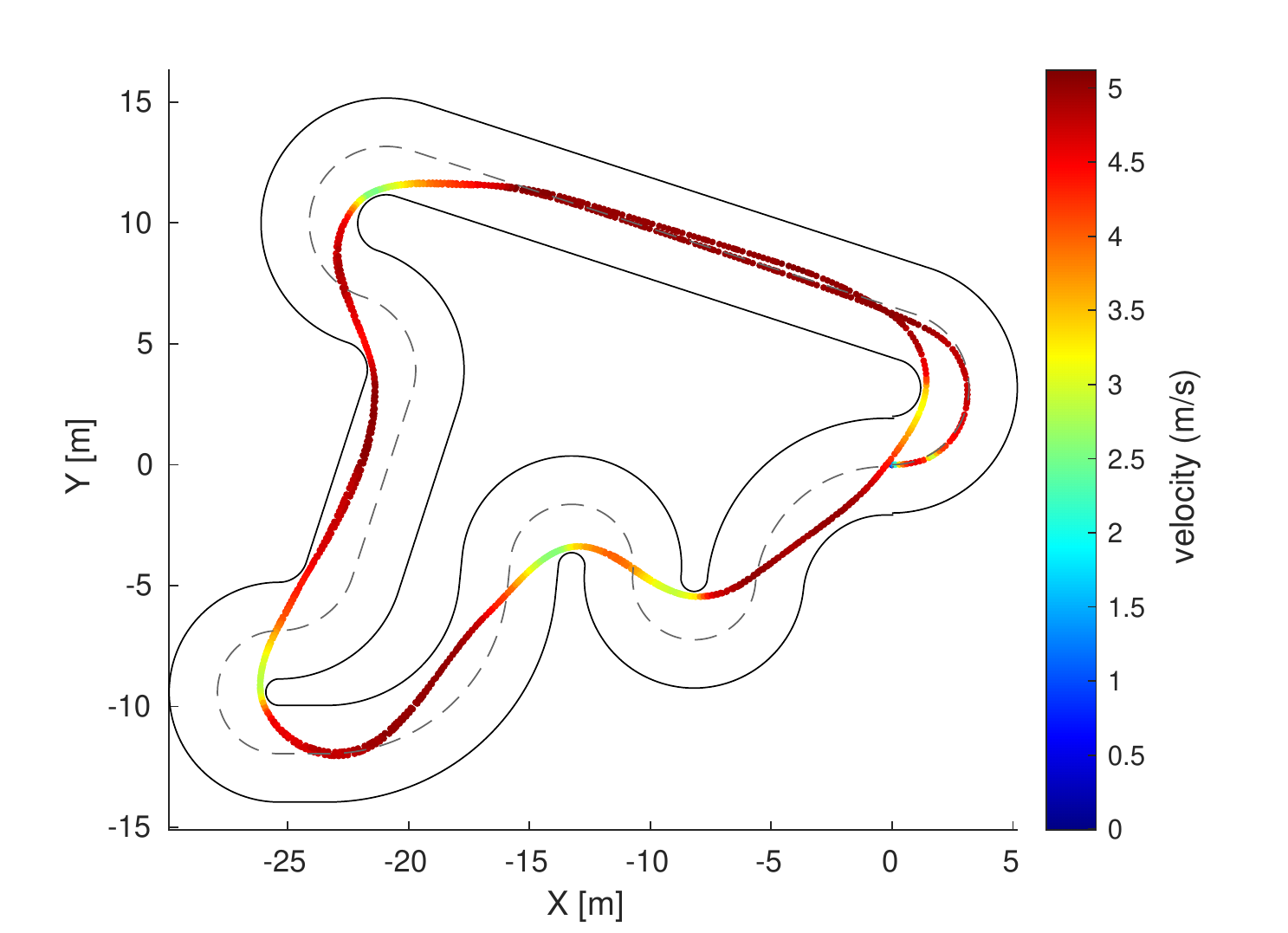}};    
        \end{tikzpicture}
		\label{subfig:senario3NoObstacles}
    \end{subfigure}
    \vspace{-0.4cm}
    \\
    \hspace{-0.64cm}
    \begin{subfigure}{0.33\columnwidth}
        \centering
        \vspace{-0.275cm}
        \hspace{-0.5cm}
        \begin{tikzpicture}
        \node[inner sep=0pt] (model) at (0, 0) {\adjincludegraphics[width=1.9\textwidth, keepaspectratio, trim={{.05\width} {.0\height} {.05\width} {.07\height}}, clip]{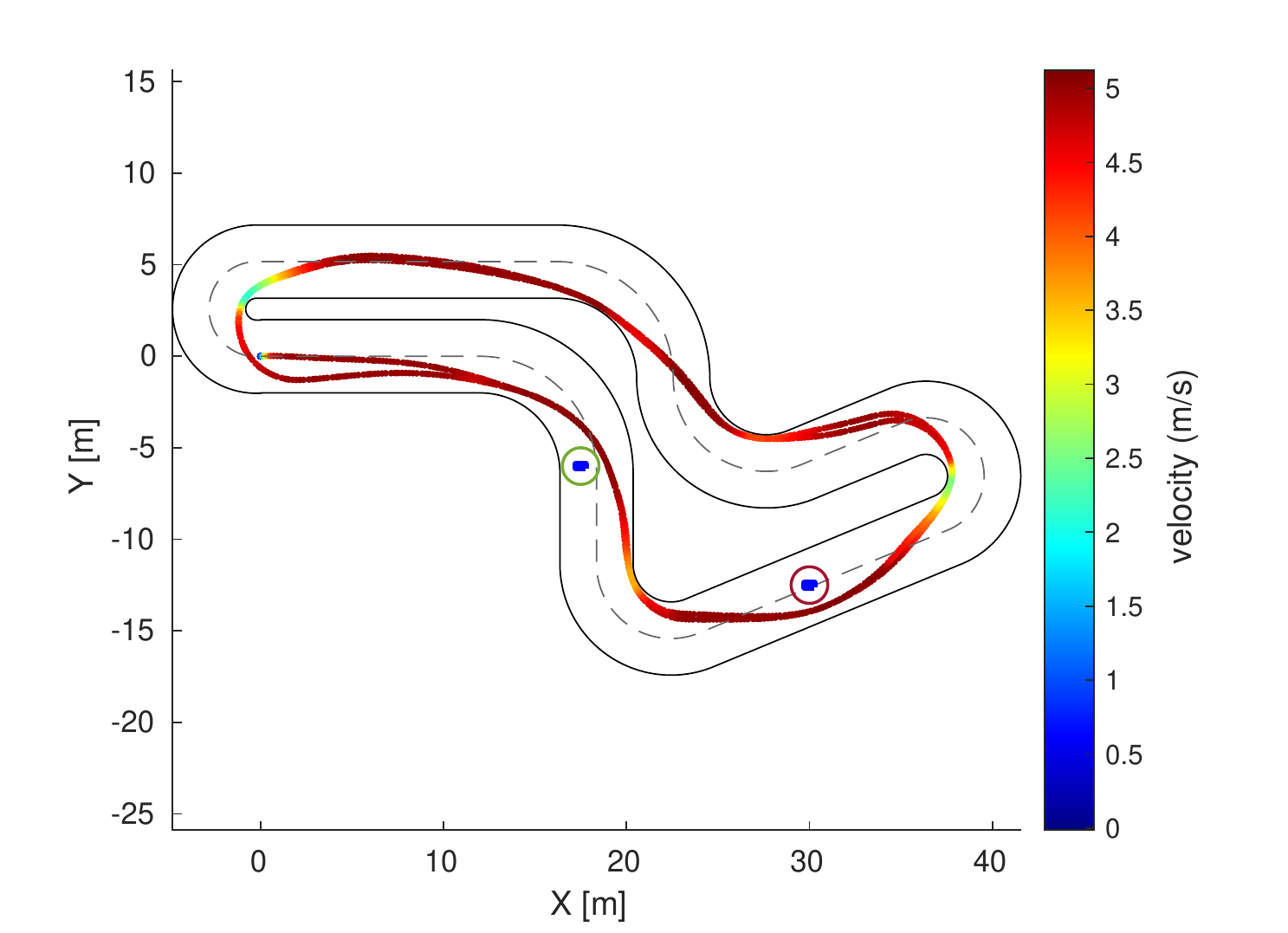}};    
        \end{tikzpicture}
		\label{subfig:senario1WithObstacles}
    \end{subfigure}
    \hspace{2.9cm}
    \begin{subfigure}{0.33\columnwidth}
        \centering
        \vspace{-0.275cm}
        \hspace{-0.5cm}
        \begin{tikzpicture}
        \node[inner sep=0pt] (model) at (0, 0) {\adjincludegraphics[width=1.9\textwidth, keepaspectratio, trim={{.05\width} {.0\height} {.05\width} {.07\height}}, clip]{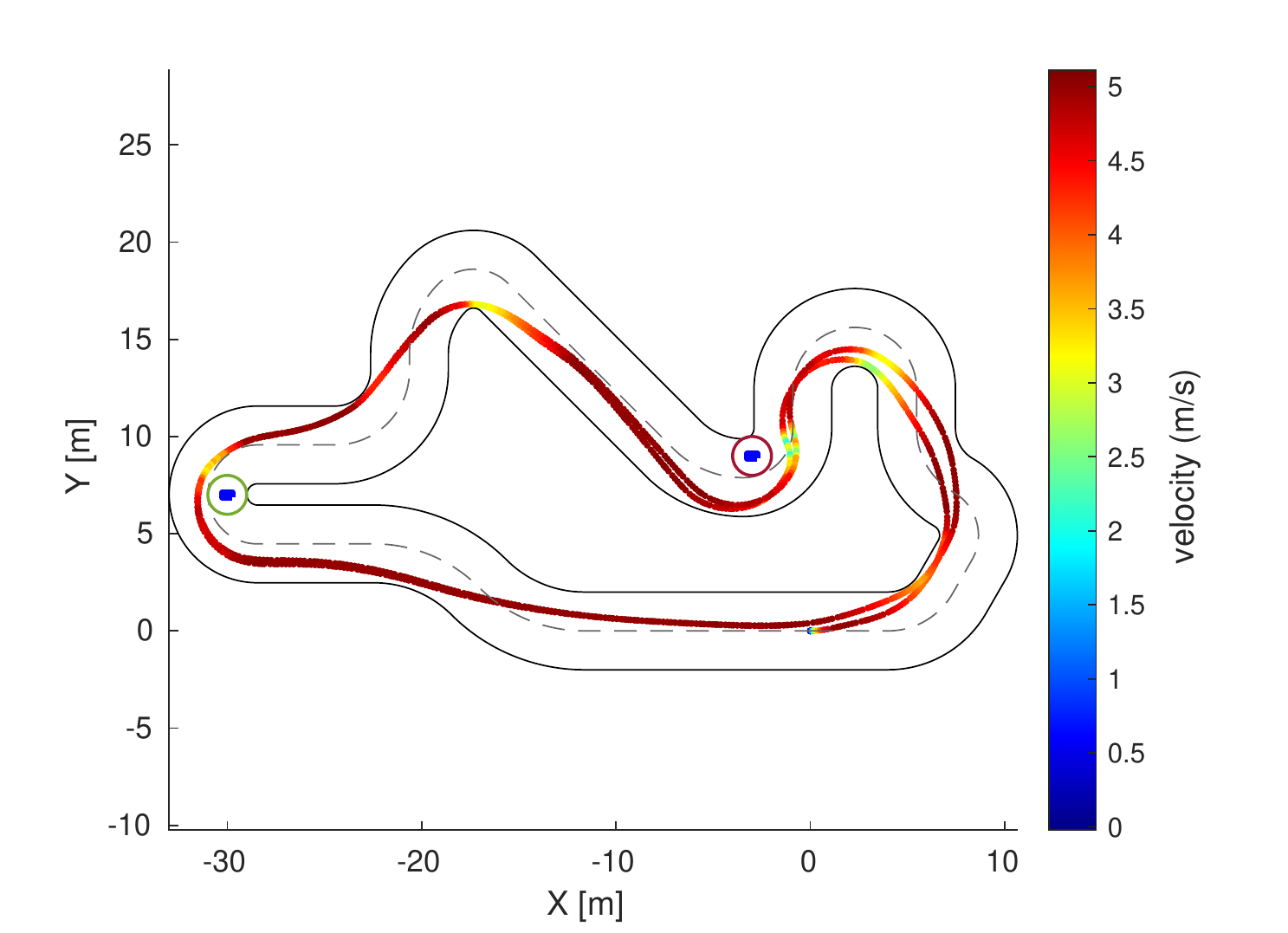}};    
        \end{tikzpicture}
		\label{subfig:senario2WithObstacles}
    \end{subfigure}
    \hspace{2.9cm}
    \begin{subfigure}{0.33\columnwidth}
        \centering
        \vspace{-0.275cm}
        \hspace{-0.5cm}
        \begin{tikzpicture}
        \node[inner sep=0pt] (model) at (0, 0) {\adjincludegraphics[width=1.9\textwidth, keepaspectratio, trim={{.065\width} {.0\height} {.07\width} {.07\height}}, clip]{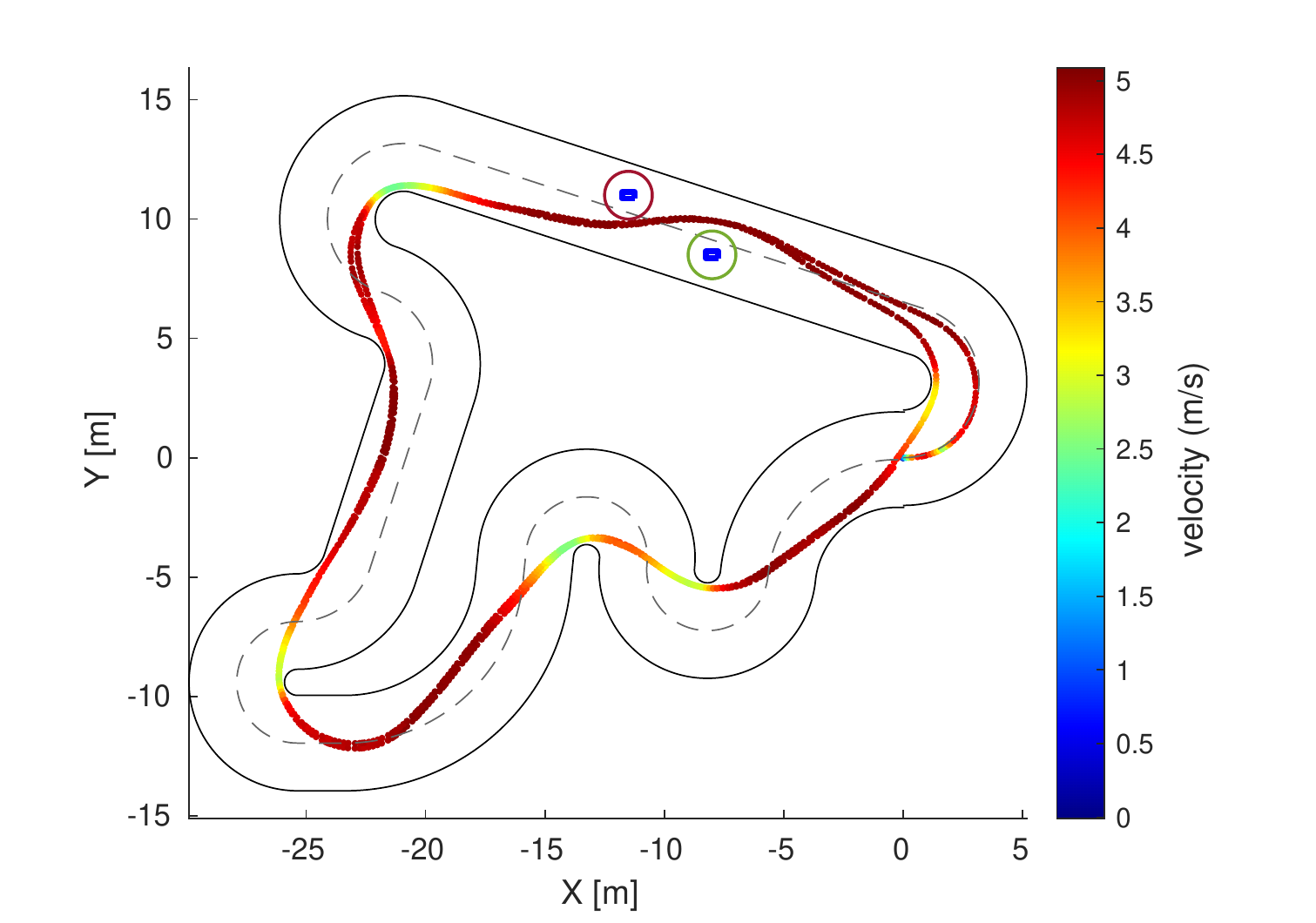}};    
        \end{tikzpicture}
		\label{subfig:senario3WithObstacles}
    \end{subfigure}
    \vspace{-1.75em}
    \caption{Simulation tracks. The color gradient indicates the car's velocity. 
    Track scenarios with obstacles are at the bottom.}
    \label{fig:scenariosWithObstacles}
\end{figure*}

Three different tracks in which the vehicle run in counter-clockwise direction were considered. Figure~\ref{fig:scenariosWithObstacles} shows the driven trajectories in the world frame along with the velocity profile encoded in gradient colors for each of them. In all simulated scenarios, the car's velocity touches the actuation limits imposed by the vehicle dynamics, maintaining high values for most of the track. Minimum velocity values can be seen in the most demanding stretches, where the minimum value is just under $\SI{3}{\meter\per\second}$. It can be seen how the car slows down in sharper turns and accelerates at the exit of the turns.

\begin{figure*}[h]
    \centering
    \hspace{-3.25cm}
    \begin{subfigure}{0.33\columnwidth}
        \centering
        \begin{tikzpicture}
	\begin{axis}[%
	width=1.9119in,%
	height=0.9183in,%
	at={(0.758in,0.481in)},%
	scale only axis,%
	xmin=0,%
	xmax=40,%
	ymax=1.1,%
	ymin=0,%
	xmajorgrids,%
	ymajorgrids,%
	xticklabels={,,},%
	extra y tick style={grid=none},
	extra y ticks={0,1},
	extra y tick labels={$\hspace{-1.0cm}\underline{\tilde{d}}$, $\hspace{-1.0cm}\bar{\tilde{d}}$},
	ylabel style={yshift=-0.4215cm}, 
	xlabel style={yshift=0.115cm}, 
	ylabel={[1]},%
	axis background/.style={fill=white},%
	legend style={at={(0.3,0.3055)},anchor=north,legend cell 
		align=left,draw=none,legend columns=-1,align=left,draw=white!15!black}
	]
	\addplot [color=blue, dotted, line width=0.75pt]
	   file{matlabPlots/control/tilde_d_1.txt};%
	\addplot [color=red, dashed, line width=0.75pt]
	   file{matlabPlots/control/tilde_d_obs_1.txt};%
	\draw [thick, dashed] (axis cs:{0.00,1}) -- (axis cs:{40.00,1});
    \draw [thick, dashed] (axis cs:{0.00,0}) -- (axis cs:{40.00,0});
	\legend{$\tilde{d}$, $\tilde{d}_\mathrm{obs}$};%
	\end{axis}
\end{tikzpicture}
        \label{subfig:scenario1ControlInputs-TildeD}
    \end{subfigure}
    \hspace{3.0cm}
    \begin{subfigure}{0.33\columnwidth}
        \centering
        \begin{tikzpicture}
	\begin{axis}[%
	width=1.9119in,%
	height=0.9183in,%
	at={(0.758in,0.481in)},%
	scale only axis,%
	xmin=0,%
	xmax=40,%
	ymax=1.1,%
	ymin=0,%
	xmajorgrids,%
	ymajorgrids,%
	xticklabels={,,},%
	ylabel style={yshift=-0.4215cm}, 
	xlabel style={yshift=0.115cm}, 
	axis background/.style={fill=white},%
	legend style={at={(0.45,0.355)},anchor=north,legend cell 
		align=left,draw=none,legend columns=-1,align=left,draw=white!15!black}
	]
	\addplot [color=blue, dotted, line width=0.75pt]
	   file{matlabPlots/control/tilde_d_2.txt};%
	\addplot [color=red, dashed, line width=0.75pt]
	   file{matlabPlots/control/tilde_d_obs_2.txt};%
	\draw [thick, dashed] (axis cs:{0.00,1}) -- (axis cs:{40.00,1});
    \draw [thick, dashed] (axis cs:{0.00,0}) -- (axis cs:{40.00,0});
	\legend{$\tilde{d}$, $\tilde{d}_\mathrm{obs}$};%
	\end{axis}
\end{tikzpicture}
        \label{subfig:scenario2ControlInputs-TildeD}
    \end{subfigure}
    \hspace{2.6cm}
    \begin{subfigure}{0.33\columnwidth}
        \centering
        \begin{tikzpicture}
	\begin{axis}[%
	width=1.9119in,%
	height=0.9183in,%
	at={(0.758in,0.481in)},%
	scale only axis,%
	xmin=0,%
	xmax=40,%
	ymax=1.1,%
	ymin=0,%
	xmajorgrids,%
	ymajorgrids,%
	xticklabels={,,},%
	ylabel style={yshift=-0.4215cm}, 
	xlabel style={yshift=0.115cm}, 
	axis background/.style={fill=white},%
	legend style={at={(0.5,0.305)},anchor=north,legend cell 
		align=left,draw=none,legend columns=-1,align=left,draw=white!15!black}
	]
	\addplot [color=blue, dotted, line width=0.75pt]
	   file{matlabPlots/control/tilde_d_3.txt};%
	\addplot [color=red, dashed, line width=0.75pt]
	   file{matlabPlots/control/tilde_d_obs_3.txt};%
	\draw [thick, dashed] (axis cs:{0.00,1}) -- (axis cs:{40.00,1});
    \draw [thick, dashed] (axis cs:{0.00,0}) -- (axis cs:{40.00,0});
	\legend{$\tilde{d}$, $\tilde{d}_\mathrm{obs}$};%
	\end{axis}
\end{tikzpicture}
        \label{subfig:scenario3ControlInputs-TildeD}
    \end{subfigure}
    \\
    \vspace{-0.45cm}
    \hspace{-3.39cm}
    \begin{subfigure}{0.33\columnwidth}
        \centering
        \begin{tikzpicture}
	\begin{axis}[%
	width=1.9119in,%
	height=0.9183in,%
	at={(0.758in,0.481in)},%
	scale only axis,%
	xmin=0,%
	xmax=40,%
	ymax=35,%
	ymin=-35,%
	xmajorgrids,%
	ymajorgrids,%
	extra y tick style={grid=none},
	extra y ticks={-30,30},
	extra y tick labels={$\hspace{-1.0cm}\underline{\delta}$, $\hspace{-1.0cm}\bar{\delta}$},
	ylabel style={yshift=-0.415cm}, 
	xlabel style={yshift=0.115cm}, 
	ylabel={[$\si{\degree}$]},%
	xlabel={Time [$\si{\second}$]},%
	axis background/.style={fill=white},%
	legend style={at={(0.4,0.305)},anchor=north,legend cell 
		align=left,draw=none,legend columns=-1,align=left,draw=white!15!black}
	]
	\addplot [color=blue, dotted, line width=0.75pt]
	   file{matlabPlots/control/delta_1.txt};%
	\addplot [color=red, dashed, line width=0.75pt]
	   file{matlabPlots/control/delta_obs_1.txt};%
	%
	\draw [thick, dashed] (axis cs:{0.00,30}) -- (axis cs:{40.00,30});
    \draw [thick, dashed] (axis cs:{0.00,-30}) -- (axis cs:{40.00,-30});
	\legend{$\delta$, $\delta_\mathrm{obs}$};%
	\end{axis}
\end{tikzpicture}
        \label{subfig:scenario1ControlInputs-Delta}
    \end{subfigure}
    \hspace{2.9cm}
    \begin{subfigure}{0.33\columnwidth}
        \centering
        \begin{tikzpicture}
	\begin{axis}[%
	width=1.9119in,%
	height=0.9183in,%
	at={(0.758in,0.481in)},%
	scale only axis,%
	xmin=0,%
	xmax=40,%
	ymax=35,%
	ymin=-35,%
	xmajorgrids,%
	ymajorgrids,%
	ylabel style={yshift=-0.415cm}, 
	xlabel style={yshift=0.115cm}, 
	xlabel={Time [$\si{\second}$]},%
	axis background/.style={fill=white},%
	legend style={at={(0.445,0.2855)},anchor=north,legend cell 
		align=left,draw=none,legend columns=-1,align=left,draw=white!15!black}
	]
	\addplot [color=blue, dotted, line width=0.75pt]
	   file{matlabPlots/control/delta_2.txt};%
	\addplot [color=red, dashed, line width=0.75pt]
	   file{matlabPlots/control/delta_obs_2.txt};%
	%
	\draw [thick, dashed] (axis cs:{0.00,30}) -- (axis cs:{40.00,30});
    \draw [thick, dashed] (axis cs:{0.00,-30}) -- (axis cs:{40.00,-30});
	\legend{$\delta$, $\delta_\mathrm{obs}$};%
	\end{axis}
\end{tikzpicture}
        \label{subfig:scenario2ControlInputs-Delta}
    \end{subfigure}
    \hspace{2.6cm}
    \begin{subfigure}{0.33\columnwidth}
        \centering
        \begin{tikzpicture}
	\begin{axis}[%
	width=1.9119in,%
	height=0.9183in,%
	at={(0.758in,0.481in)},%
	scale only axis,%
	xmin=0,%
	xmax=40,%
	ymax=35,%
	ymin=-35,%
	xmajorgrids,%
	ymajorgrids,%
	ylabel style={yshift=-0.415cm}, 
	xlabel style={yshift=0.115cm}, 
	xlabel={Time [$\si{\second}$]},%
	axis background/.style={fill=white},%
	legend style={at={(0.5,0.255)},anchor=north,legend cell 
		align=left,draw=none,legend columns=-1,align=left,draw=white!15!black}
	]
	\addplot [color=blue, dotted, line width=0.75pt]
	   file{matlabPlots/control/delta_3.txt};%
	\addplot [color=red, dashed, line width=0.75pt]
	   file{matlabPlots/control/delta_obs_3.txt};%
	%
	\draw [thick, dashed] (axis cs:{0.00,30}) -- (axis cs:{40.00,30});
    \draw [thick, dashed] (axis cs:{0.00,-30}) -- (axis cs:{40.00,-30});
	\legend{$\delta$, $\delta_\mathrm{obs}$};%
	\end{axis}
\end{tikzpicture}
        \label{subfig:scenario3ControlInputs-Delta}
    \end{subfigure}
    \vspace{-2.0em}
    \caption{Control inputs $\mathbf{u}$ in cases with ($\tilde{d}_\mathrm{obs}$, $\delta_\mathrm{obs}$) and without ($\tilde{d}$, $\delta$) obstacles. The scenarios are in the same order as Fig.~\ref{fig:scenariosWithObstacles}. }
    \label{fig:scenariosControlInputs}
\end{figure*}
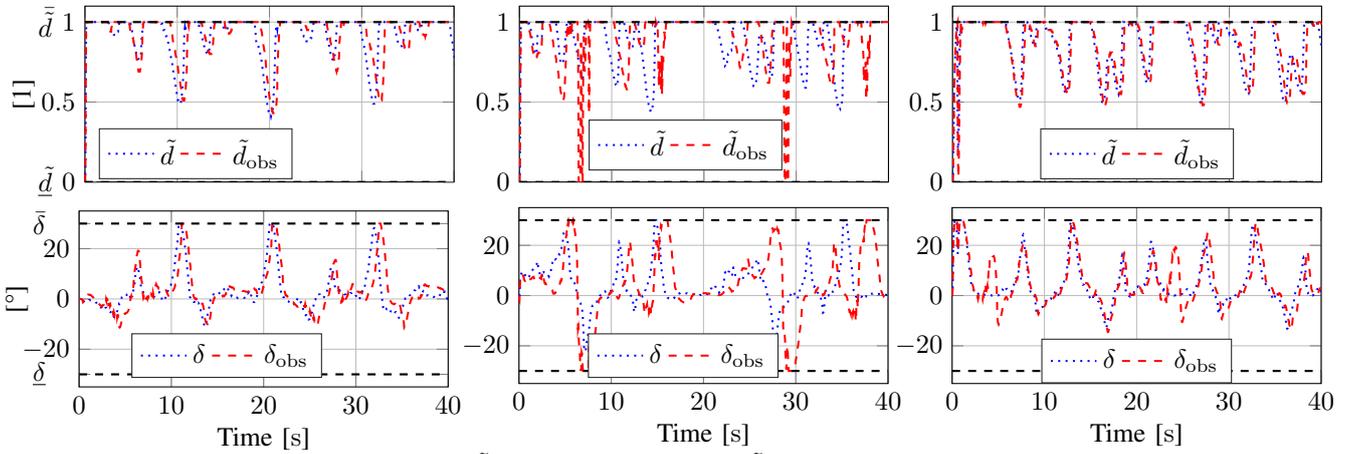

Figure~\ref{fig:scenariosWithObstacles} reports also the controller behavior in presence of obstacles (bottom graphs) showing how it adapts the car's motion to avoid collisions. 
Figure~\ref{fig:scenariosControlInputs} shows the values of the control inputs $\mathbf{u}=[\tilde{d}, \delta]^\top$ for all the scenarios. As can be seen from the graphs, the control inputs remain within the boundaries~\eqref{subeq:uBound}.


\begin{figure*}[b]
    \begin{subfigure}{0.33\columnwidth}
        \centering
        \begin{tikzpicture}
            \centering
            \node[inner sep=0pt] (model) at (0, 0) {\adjincludegraphics[width=2.2\textwidth, height=1.1\columnwidth, trim={{.0\width} {.0\height} {.0\width} {.05\height}}, clip]{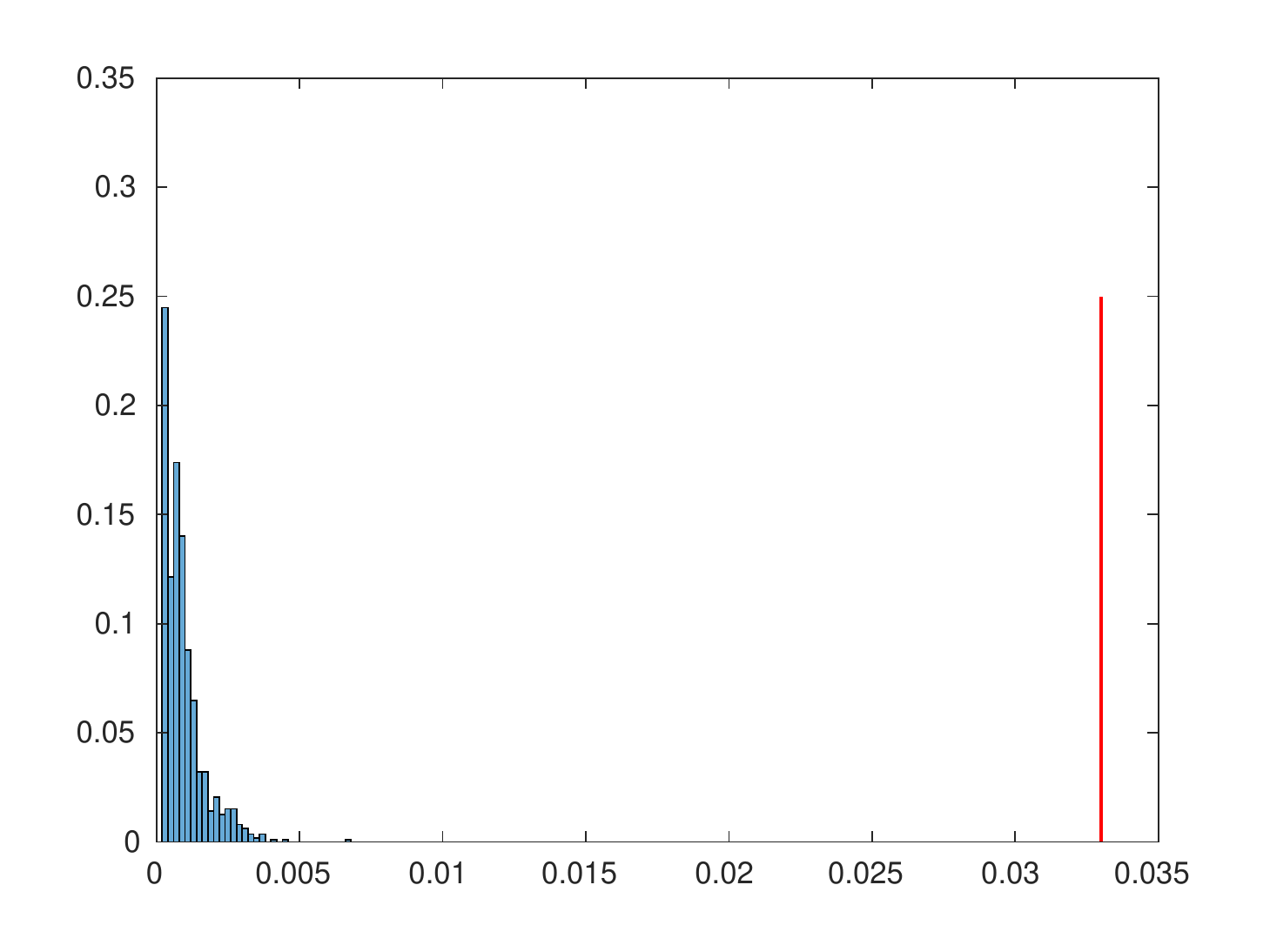}};
            \draw (-2.95,0) node[text centered, rotate=90]{\% control steps};
        \end{tikzpicture} 
		\vspace{0.3cm}
		\label{subfig:histogram1NoObstacles}
    \end{subfigure}
    \hspace{2.80cm}
    \begin{subfigure}{0.33\columnwidth}
        \centering
        \begin{tikzpicture}
            \centering
            \node[inner sep=0pt] (model) at (0, 0) {\adjincludegraphics[width=2.2\textwidth, height=1.1\columnwidth, trim={{.0\width} {.0\height} {.0\width} {.05\height}}, clip]{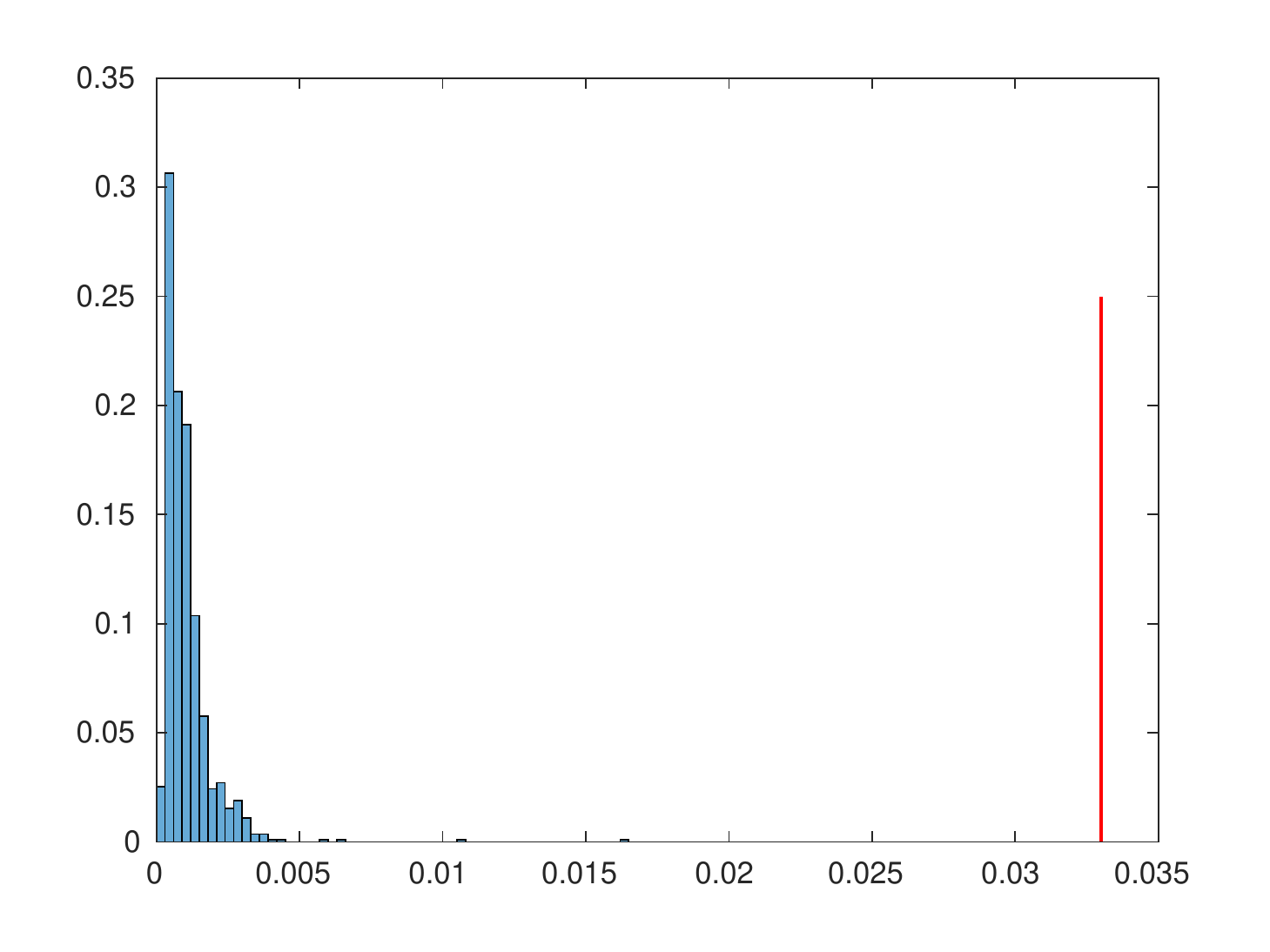}};
        \end{tikzpicture} 
		\vspace{0.3cm}
		\label{subfig:histogram2NoObstacles}
    \end{subfigure}
    \hspace{2.80cm}
    \begin{subfigure}{0.33\columnwidth}
        \centering
        \begin{tikzpicture}
            \centering
            \node[inner sep=0pt] (model) at (0, 0) {\adjincludegraphics[width=2.2\textwidth, height=1.1\columnwidth, trim={{.0\width} {.0\height} {.0\width} {.05\height}}, clip]{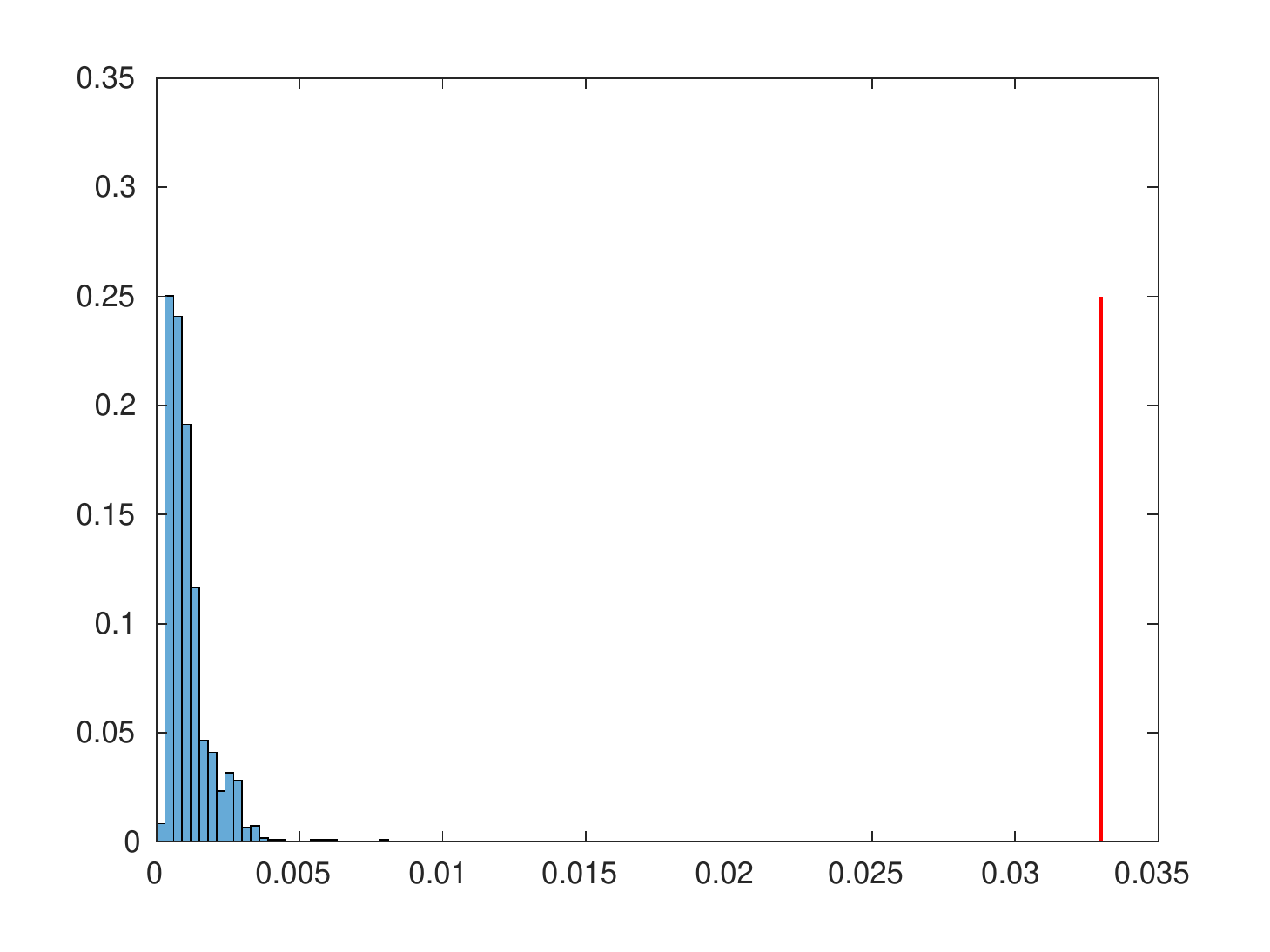}};
        \end{tikzpicture} 
		\vspace{0.3cm}
		\label{subfig:histogram3NoObstacles}
    \end{subfigure}
    \\
    \hspace{-0.6cm}
    \begin{subfigure}{0.33\columnwidth}
        \centering
        \vspace{-0.75cm}
        \begin{tikzpicture}
            \centering
            \node[inner sep=0pt] (model) at (0, 0) {\adjincludegraphics[width=2.2\textwidth, height=1.1\columnwidth, trim={{.0\width} {.0\height} {.0\width} {.05\height}}, clip]{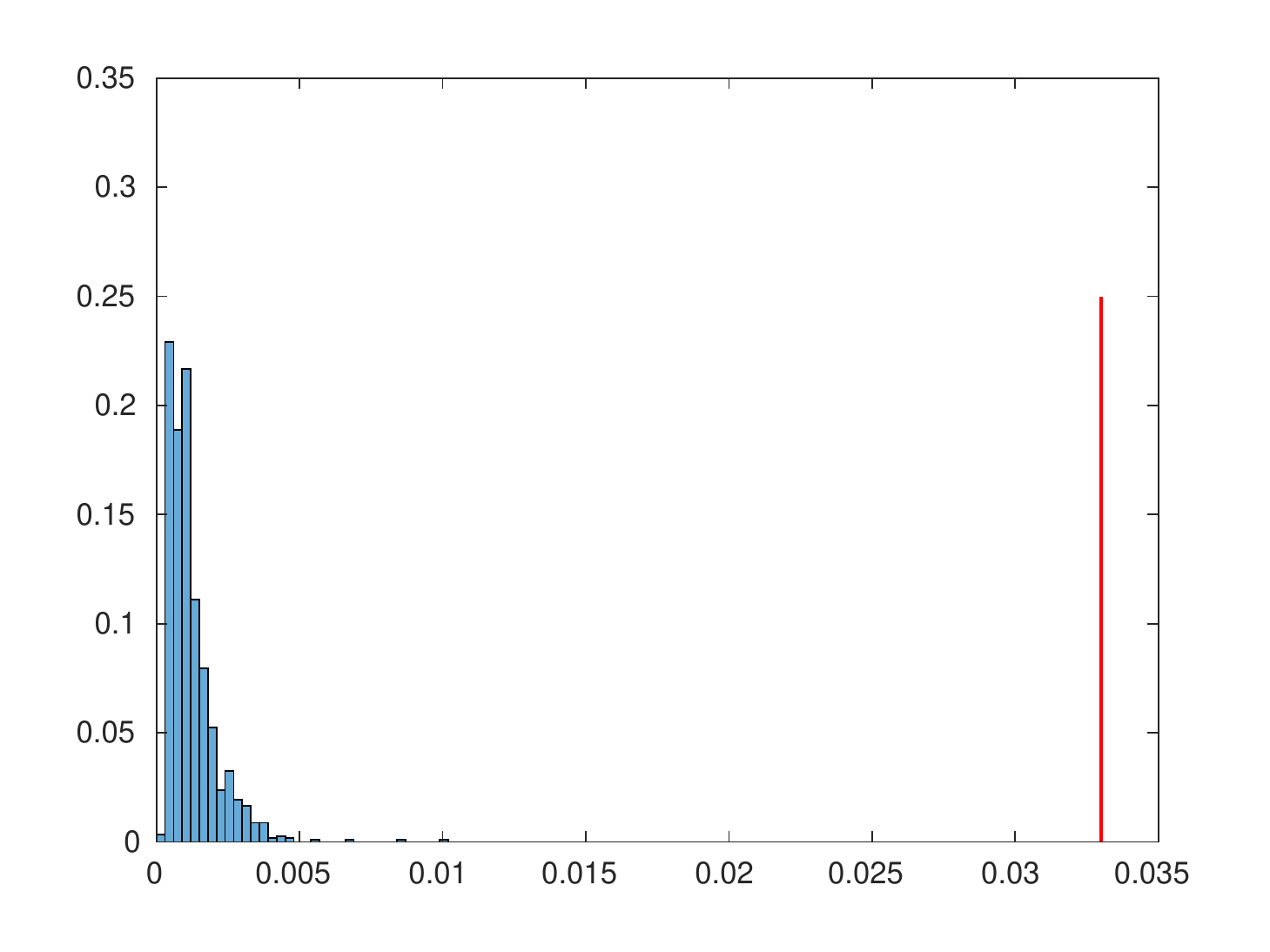}};
            \draw (0,-1.75) node[text centered]{\text{Time [s]}};
            \draw (-2.95,0) node[text centered, rotate=90]{\% control steps};
        \end{tikzpicture} 
		\label{subfig:histogram1WithObstacles}
    \end{subfigure}
    \hspace{2.80cm}
    \begin{subfigure}{0.33\columnwidth}
        \centering
        \vspace{-0.75cm}
        \begin{tikzpicture}
            \centering
            \node[inner sep=0pt] (model) at (0, 0) {\adjincludegraphics[width=2.2\textwidth, height=1.1\columnwidth, trim={{.0\width} {.0\height} {.0\width} {.05\height}}, clip]{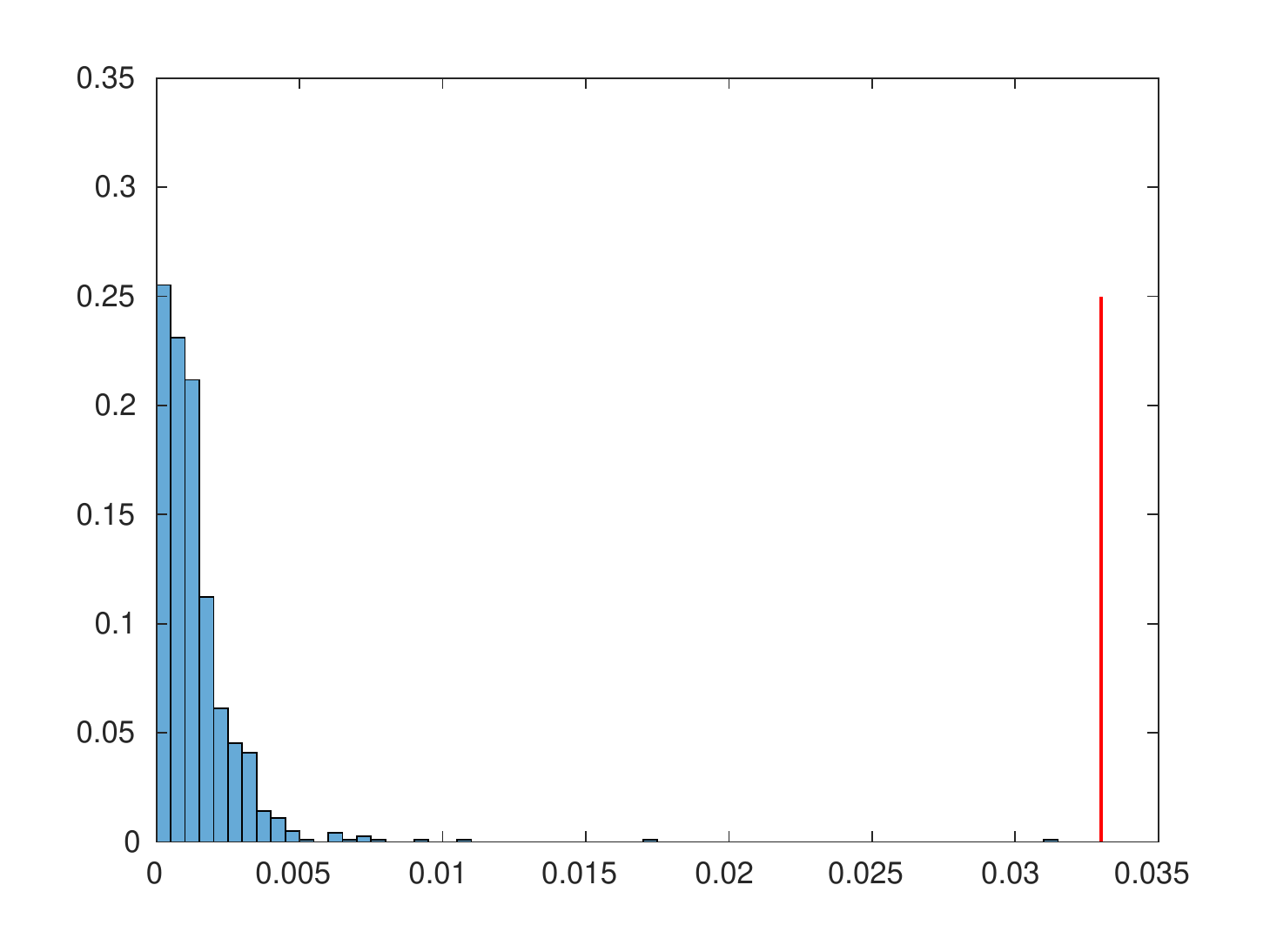}};
            \draw (0,-1.75) node[text centered]{\text{Time [s]}};
        \end{tikzpicture} 
		\label{subfig:histogram2WithObstacles}
    \end{subfigure}
    \hspace{2.80cm}
    \begin{subfigure}{0.33\columnwidth}
        \centering
        \vspace{-0.75cm}
        \begin{tikzpicture}
            \centering
            \node[inner sep=0pt] (model) at (0, 0) {\adjincludegraphics[width=2.2\textwidth, height=1.1\columnwidth, trim={{.0\width} {.0\height} {.0\width} {.05\height}}, clip]{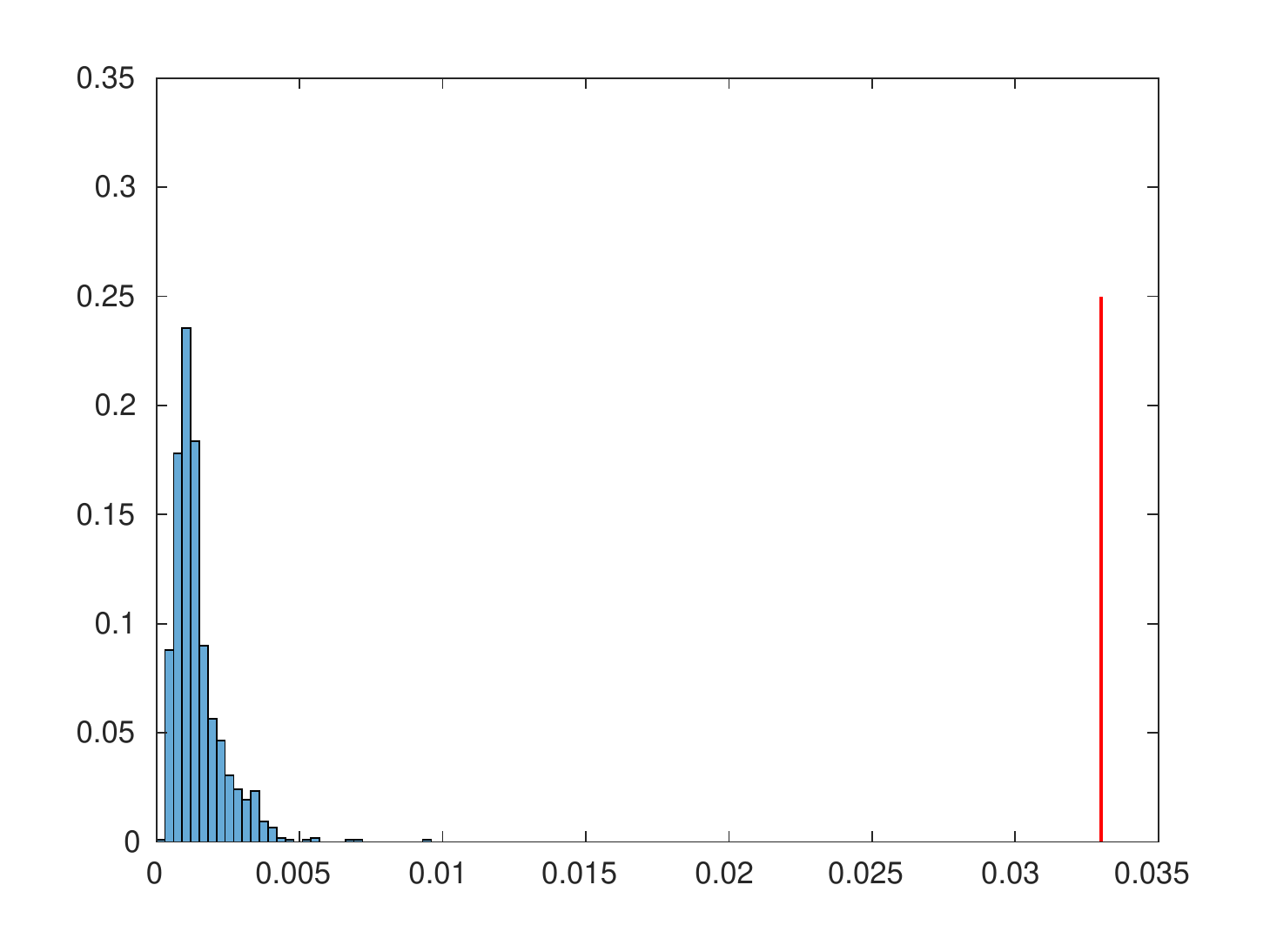}};
            \draw (0,-1.75) node[text centered]{\text{Time [s]}};
        \end{tikzpicture} 
		\label{subfig:histogram3WithObstacles}
    \end{subfigure}
    \vspace{-1.75em}
    \caption{Histogram of the computation time of the NMPC step. The order follows that of Fig.~\ref{fig:scenariosWithObstacles}. The red bar is $T_s = \SI{33}{\milli\second}$.}
    \label{fig:histogramWithObstacles}
\end{figure*}
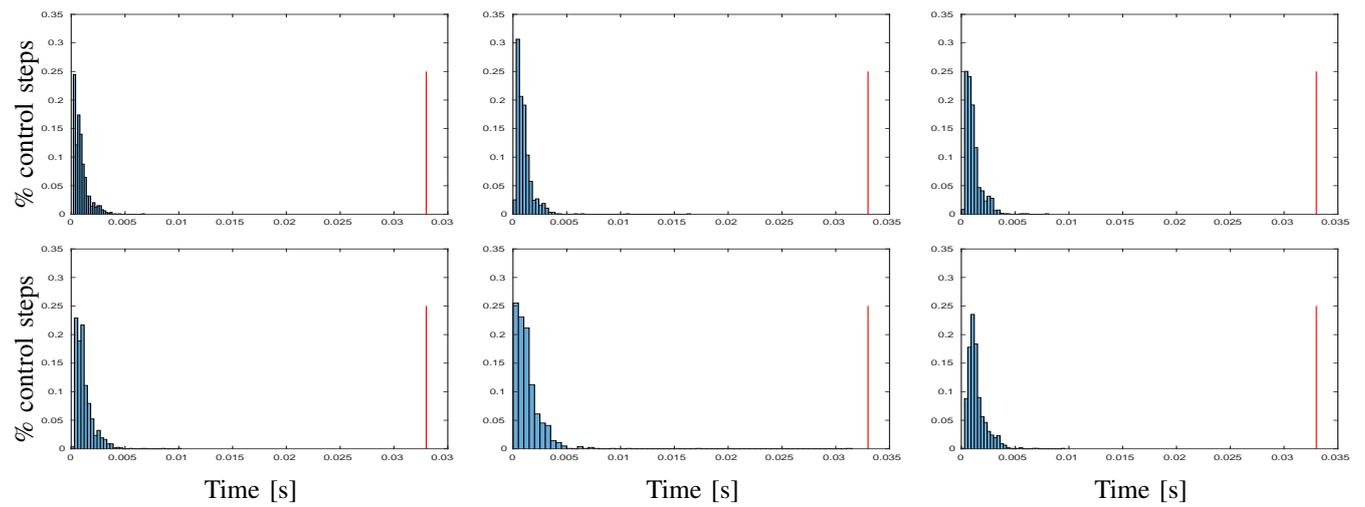

Figure~\ref{fig:histogramWithObstacles}  shows the histograms of the computation time per control step of the~\ac{NMPC} strategy. Low computation time per control step is also showcased when considering static obstacles along the track. Hence, these graphs show the capability of the proposed framework to compute the necessary control actions to drive the vehicle pushing the limits without violating the constraints. 

Table~\ref{tab:computationTime} reports the average computation time for the~\ac{NMPC}. The minimum and the maximum average computation time obtained during the test runs are $\SI{0.9}{\milli\second}$ and $\SI{1.4}{\milli\second}$, respectively, such that the sampling time of $\SI{33}{\milli\second}$ is never missed. The former is retrieved in the case without obstacles and with the shortest number of stretches, the latter refers to the last simulations where the vehicle is demanded to race on a winding track populated by obstacles.

\begin{table}[tb]
    \begin{center}
    \begin{tabular}{|l|l|l|}
    \hline
    \textbf{Average Computation Time} & \textbf{Without Obs.} & \textbf{With Obs.} \\
    \hline \hline
    \text{Scenario 1} & $\SI{0.9}{\milli\second}$ & $\SI{1.2}{\milli\second}$ \\
    \text{Scenario 2} & $\SI{1.0}{\milli\second}$ & $\SI{1.4}{\milli\second}$ \\
    \text{Scenario 3} & $\SI{1.1}{\milli\second}$ & $\SI{1.4}{\milli\second}$ \\
    \hline
    \end{tabular}
    \caption{Average computation time of the~\ac{NMPC} for different scenarios.}
    \label{tab:computationTime}
    \end{center}
\end{table}



\section{Conclusions}
\label{sec:conclusions}

In this paper, an~\ac{NMPC} strategy for autonomous racing of scale vehicles was presented. 
Numerical simulations performed in the F1/10 simulator demonstrated the validity of the proposed control strategy in a scenario quite close to real implementations. The proposed approach has shown low computation times to solve the optimization problem making it effective for complex control maneuvers, such as those in autonomous racing applications. Future work will include advanced path planning solutions to deal with uncertainties on the lane center line position and more challenging scenarios will be investigated, such as the combination of static and dynamic obstacles, in the direction of field experiments.



\bibliographystyle{IEEEtran}
\bibliography{bib_short.bib}

\end{document}